\documentclass{article}

     \PassOptionsToPackage{numbers, compress}{natbib}

     \usepackage[final]{neurips_2020}

\usepackage[utf8]{inputenc} 
\usepackage[T1]{fontenc}    
\usepackage{hyperref}       
\usepackage{url}            
\usepackage{booktabs}       
\usepackage{amsfonts}       
\usepackage{nicefrac}       
\usepackage{microtype}      
\usepackage{url}
\usepackage{multirow}
\usepackage[]{bm}
\usepackage{booktabs}
\usepackage{rotating}
\usepackage{amsfonts}
\usepackage{graphicx,wrapfig}
\usepackage{subcaption}
\usepackage{mathtools}
\usepackage{amsmath,amssymb,amsfonts}
\usepackage{color,soul}
\usepackage{floatflt}

\title{Towards Maximizing the Representation Gap between In-Domain \& Out-of-Distribution Examples}
\author{%
  Jay Nandy ~~~~~ Wynne Hsu ~~~~~ Mong Li Lee\\
  National University of Singapore\\
  \texttt{\{jaynandy,whsu,leeml\}@comp.nus.edu.sg} \\
}

\begin{document}

\maketitle

\begin{abstract}

Among existing uncertainty estimation approaches, Dirichlet Prior Network (DPN) distinctly models different predictive uncertainty types.
However, for in-domain examples with high data uncertainties among multiple classes, even a DPN model often produces indistinguishable representations from the out-of-distribution (OOD) examples, compromising their OOD detection performance.
We address this shortcoming by proposing a novel loss function for DPN to maximize the \textit{representation gap} between in-domain and OOD examples.
Experimental results demonstrate that our proposed approach consistently improves OOD detection performance.
\end{abstract}

\section{Introduction}
Deep neural network (DNN) based models have achieved impeccable success to address various real-world tasks \cite{vgg16,speech_1,medical_1}. 
However, when these intelligent systems fail, they do not provide any explanation or warning.
Predictive uncertainty estimation has emerged as an important research direction
to inform users about possible wrong predictions and allow users to react in an informative manner, thus improving their reliability. 

Predictive uncertainties of DNNs can come from three different sources: \textit{model uncertainty, data uncertainty,} and \textit{distributional uncertainty} \cite{gal_thesis_2016,dpn_nips_2018}.
Model uncertainty or epistemic uncertainty captures the uncertainty in estimating the model parameters, conditioning on training data \cite{gal_thesis_2016}.
Data uncertainty (or aleatoric uncertainty) arises from the natural complexities of the underlying distribution, such as class overlap, label noise, homoscedastic and heteroscedastic noise, etc \cite{gal_thesis_2016}.
Distributional uncertainty or dataset shift arises due to the distributional mismatch between the training and test examples, that is, the   test data is \textit{out-of-distribution (OOD)} \cite{data_shift_2009,dpn_nips_2018}.

It is useful to determine the sources of predictive uncertainties.
In active learning, distributional uncertainty indicates that the classifier requires additional data for training.
For real-world applications, where the cost of errors are high, such as in autonomous vehicles \cite{intro_self_drive_2016}, medical diagnosis \cite{intro_medical_2017}, financial, and legal fields \cite{intro_finance_2018}, the source of uncertainty can allow manual intervention in an informed way.

Notable progress has been made for predictive uncertainty estimation.
Bayesian neural network-based models conflate the distributional uncertainty through model uncertainty \cite{BayesianNN_icml_2015,gal_thesis_2016,mcDrop_icml_gal16,mnfs_icml_2017,ensemble_nips_2017}.
However, since the true posterior for their model parameters are intractable, their success depend on the nature of approximations. 
In contrast, non-Bayesian approaches can explicitly train the network in a multi-task fashion, incorporating both in-domain and OOD examples to produce sharp and uniform categorical predictions respectively \cite{gan_ood_2018,oe_iclr_2019}.
However, these approaches cannot robustly determine the source of predictive uncertainty \cite{dpn_nips_2018}.
In particular, the presence of high data uncertainty among multiple classes leads them to produce uniform categorical predictions for in-domain examples, often making them indistinguishable from the OOD examples.

Dirichlet Prior Network (DPN) separately models different uncertainty types by producing sharp uni-modal Dirichlet distributions for in-domain examples, and flat Dirichlet distributions for OOD examples \cite{dpn_nips_2018,dpn2_nips_2019}.
It uses a loss function that explicitly incorporates Kullback-Leibler (KL)-divergence between the model output and a target Dirichlet with a pre-specified precision value.
However, we show that for in-domain examples with high data uncertainties, their proposed loss function distributes the target precision values among the overlapping classes, leading to much flatter distributions.
Hence, it often produces indistinguishable representations for such in-domain misclassified examples and OOD examples, compromising the OOD detection performance.

In this work, we propose an alternative approach for a DPN classifier that produces \textit{sharp, multi-modal} Dirichlet distributions for OOD examples to maximize their \textit{representation gap} from in-domain examples. 
We design a new loss function that separately models the mean and the precision of the output Dirichlet distributions by introducing a novel \textit{explicit precision regularizer} along with the cross-entropy loss.
Experimental results on several benchmark datasets demonstrate that our proposed approach achieves the best OOD detection performance.
\section{Related Work}
In the Bayesian neural network, the predictive uncertainty of a classification model is expressed in terms of data and model uncertainty \cite{gal_thesis_2016}.
Let $\mathcal{D}_{in} = \{{\bm x_i}, y_i\}_{i=1}^N \sim P_{in}({\bm x}, y)$ where  $\bm{x}$ and $y$ denotes the images and their corresponding class-labels, sampled from an underlying probability distribution $P_{in}({\bm x}, y)$.
Given an input ${\bm x^*}$, the data uncertainty, $p(\omega_c| {\bm x^*}, \bm \theta)$ is the posterior distribution over class labels given the model parameters $\bm \theta$, while the model uncertainty, $p({\bm \theta} | \mathcal{D}_{in})$ is the posterior distribution over parameters given the data, $\mathcal{D}_{in}$.
Hence, the predictive uncertainty is given as:
\begin{equation}
\label{eq:bayesian_uncertain}
\small
p(\omega_c| {\bm x^*}, \mathcal{D}_{in}) 
= \int p(\omega_c| {\bm x^*}, \bm \theta) ~~ p({\bm \theta} | \mathcal{D}_{in})~~ d{\bm \theta}
\end{equation}
where $\omega_c$ is the representation for class $c$. 
We use the standard abbreviation for 
$\small p(y= \omega_c| {\bm x^*}, \mathcal{D}_{in})$ as $p(\omega_c| {\bm x^*}, \mathcal{D}_{in})$.

\normalsize{}
However, the true posterior of $p({\bm \theta} | \mathcal{D}_{in})$ is intractable. 
Hence, we need approximation such as Monte-Carlo dropout (MCDP) \cite{mcDrop_icml_gal16}, Langevin Dynamics \cite{langevin_icml_2011}, explicit ensembling \cite{ensemble_nips_2017}: 
\small{$p(\omega_c| {\bm x^*}, \mathcal{D}_{in}) 
\approx \frac{1}{M} \sum_{m=1}^{M} p(\omega_c| {\bm x^*}, {\bm \theta}^{(m)})$.}
%
%
\normalsize{}
where, ${\bm \theta}^{(m)} \sim q({\bm \theta})$ is sampled from an explicit or implicit variational approximation, $q({\bm \theta})$ of the true posterior $p({\bm \theta} | \mathcal{D}_{in})$.
Each $p(\omega_c| {\bm x^*}, \bm \theta^{(m)})$ represents a categorical distribution, ${\bm \mu} = [\mu_1, \cdots, \mu_K] = [p(y=\omega_1), \cdots, p(y=\omega_K)]$ over class labels, given $\bm{x^*}$. 
Hence, the ensemble can be visualized as a collection of points on the probability simplex.
For a confident prediction, it should be appeared sharply in one corner of the simplex.
For an OOD example, it should be spread uniformly.
We can determine the source of uncertainty in terms of the model uncertainty by measuring their spread.
However, producing an ensemble distribution is computationally expensive. 
Further, it is difficult to control the desired behavior in practice \cite{dpn2_nips_2019}.
Furthermore, for standard DNN models, with millions of parameters, it is even harder to find an appropriate prior distribution and the inference scheme to estimate the posterior distribution of the model.

Few recent works, such as Dirichlet prior network (DPN) \cite{dpn_nips_2018,dpn2_nips_2019}, evidential deep learning (EDL) \cite{edl_nips_2018} etc, attempt to emulate this behavior by placing a Dirichlet distribution as a prior, over the predictive categorical distribution.
In particular, DPN framework \cite{dpn_nips_2018,dpn2_nips_2019} significantly improves the OOD detection performance by explicitly incorporating OOD training examples, $\mathcal{D}_{out}$, as we elaborate in Section \ref{sec:dpn}.

Non-Bayesian frameworks derive their measure of uncertainties using their predictive posteriors obtained from DNNs.
Several  works demonstrate that tweaking the input images using adversarial perturbations \cite{fgsm_2014} can enhance the performance of a DNN for OOD detection   \cite{odin_iclr_2018,mahalanobis_nips_18}.
However, these approaches are sensitive to the tuning of parameters for each OOD distribution and difficult to apply for real-world applications.
DeVries \& Taylor (2018) \cite{conf_branch_2018} propose an auxiliary confidence estimation branch to derive OOD scores.
Shalev et al. (2018) \cite{semantic_nips_18} use multiple semantic dense representations as the target label to train the OOD detection network.
The works in \cite{gan_ood_2018,oe_iclr_2019}  also introduce multi-task loss, incorporating OOD data for training.
Hein et al. (2019) \cite{relu_problem_cvpr_19} show that ReLU-networks lead to over-confident predictions even for samples that are far away from the in-domain distributions and propose methods to mitigate this problem \cite{relu_problem_cvpr_19,ccu_iclr_2020,provable2_arxiv_2020}.
While these models can identify the total predictive uncertainties, they cannot robustly determine whether the source of uncertainty is due to an in-domain data misclassification or an OOD example.

\section{Dirichlet Prior Network}
\label{sec:dpn}

\begin{wrapfigure}{r}{4.5cm}
\centering
\vspace{-1.0em}
\caption{Desired behavior of DPN classifiers to indicate different predictive uncertainty types.}
\vspace{-0.5em}
\begin{subfigure}[t]{0.45\linewidth}
\centering
\captionsetup{justification=centering}
\includegraphics[width=1\linewidth, height=35pt]{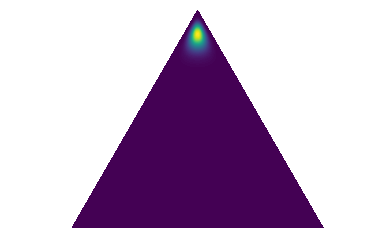}
\caption{\small Confident}
\label{fig:dirichlet_2(a)}
\end{subfigure}
\begin{subfigure}[t]{0.45\linewidth}
\centering    
\captionsetup{justification=centering}
\includegraphics[width=1\linewidth, height=35pt]{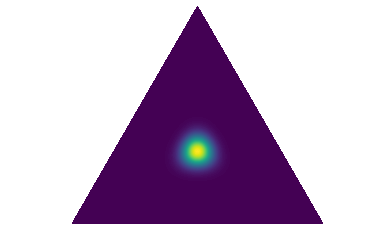}
\caption{\small Data}
\label{fig:dirichlet_2(b)}
\end{subfigure}

\begin{subfigure}[t]{0.49\linewidth}
\centering
\captionsetup{justification=centering}
\includegraphics[width=0.9\linewidth, height=35pt]{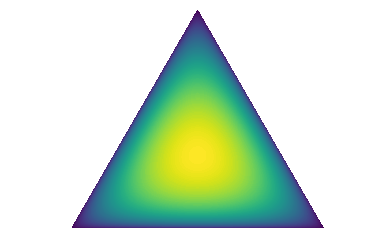}
\caption{\small Distributional \\ (Existing)}
\label{fig:dirichlet_2(c)}
\end{subfigure}
\begin{subfigure}[t]{0.49\linewidth}
\centering
\captionsetup{justification=centering}
\includegraphics[width=0.9\linewidth, height=35pt]{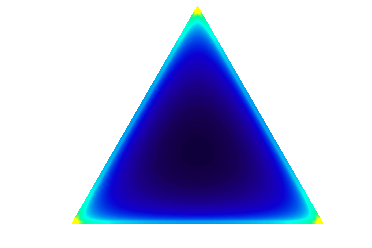}
\caption{\small Distributional \\ (Proposed)}
\label{fig:dirichlet_2(d)}
\end{subfigure}
\vspace{-0.9em}
\label{fig:dirichlet_2}
\end{wrapfigure}

A DPN classification model directly parametrizes a Dirichlet distribution as a prior to the predictive categorical distribution over the probability simplex \cite{dpn_nips_2018,dpn2_nips_2019}.
It attempts to produce a sharp Dirichlet in a corner when it is confident in its predictions for in-domain examples (Fig \ref{fig:dirichlet_2(a)}). 
For in-domain examples with high data uncertainty, it attempts to produce a sharp distribution in the middle of the simplex (Fig \ref{fig:dirichlet_2(b)}).
Note that, the probability densities for both Dirichlet distributions in Fig \ref{fig:dirichlet_2(a)} and Fig \ref{fig:dirichlet_2(b)} are concentrated in a single mode.

For OOD examples,  an existing DPN produces a flat Dirichlet distribution to indicate high-order distributional uncertainty  (see Fig \ref{fig:dirichlet_2(c)}).
However,  we demonstrate that in the case of high data uncertainty among multiple classes, an existing DPN model also produces flatter Dirichlet distributions, leading to indistinguishable representations from OOD examples (see  Section \ref{sec:proposed}).
Hence, we propose to produce sharp multi-modal Dirichlet distributions for OOD examples to increase their \textit{``representation gap"} from in-domain examples, and improve the OOD detection performance (see Fig \ref{fig:dirichlet_2(d)}).
Note that, compared to Fig \ref{fig:dirichlet_2(a)} or Fig \ref{fig:dirichlet_2(b)}, the probability densities of both Dirichlet distributions in Fig \ref{fig:dirichlet_2(c)} and Fig \ref{fig:dirichlet_2(d)} are more scattered over the simplex.
We can compute this \textit{``diversity"} using measures, such as \textit{``mutual information (MI)"}, to detect the OOD examples \cite{dpn_nips_2018}.
The predictive uncertainty of a DPN is given as:
\begin{equation} \small
\label{eq:dpn}
p(\omega_c| \bm{x^*}, \mathcal{D}) = \int \int p(\omega_c|{\bm\mu})p(\bm{\mu} | \bm{x^*}, {\bm \theta}) p({\bm \theta} | \mathcal{D})d{\bm \mu} ~d{\bm \theta}
\end{equation}
where $\mathcal{D}$ denotes the training examples comprising both in-domain, $\mathcal{D}_{in}$, and OOD  examples, $\mathcal{D}_{out}$.

Here, the data uncertainty, $p(\omega_c|{\bm\mu})$ is represented by point-estimate categorical distribution $\bm{\mu}$, while the distributional uncertainty is represented by using the distribution over the predictive categorical i.e. $p(\bm{\mu} | \bm{x^*}, \theta)$.
A high model uncertainty, $p(\bm \theta | \mathcal{D})$ would induce a high variation in distributional uncertainty, leading to larger data uncertainty.

DPN is consistent with existing approaches where an additional term is incorporated for distributional uncertainty.
Marginalization of $\bm \mu$ in Eqn. \ref{eq:dpn} produces Eqn. \ref{eq:bayesian_uncertain}.
Further, marginalizing ${\bm \theta}$ produces the expected estimation of data and distributional uncertainty given model uncertainty, i.e,
\begin{equation} \small
p(\omega_c| \bm{x^*}, \mathcal{D}) 
= \int p(\omega_c|{\bm\mu}) \Big[ \int p(\bm{\mu} | \bm{x^*}, \bm \theta) p(\bm \theta | \mathcal{D}) d{\bm \theta} \Big] d{\bm \mu}
= \int p(\omega_c|{\bm\mu})p(\bm{\mu} | \bm{x^*}, \mathcal{D}) d{\bm \mu}
\end{equation}
However, as in Eqn. \ref{eq:bayesian_uncertain}, marginalization of $\bm \theta$ is not tractable. 
Hence, as before, we can introduce approximation techniques for $p(\bm \theta | \mathcal{D})$ to measure model uncertainty \cite{mcDrop_icml_gal16,ensemble_nips_2017}.
However, the model uncertainty is \textit{reducible} given large amount of training data \cite{gal_thesis_2016}. 
Here, we only focus on the uncertainties introduced by the input examples and assume a dirac-delta approximation $\hat{\bm \theta}$ for ${\bm \theta}$ for the DPN models \cite{dpn_nips_2018}:
\small
$p({\bm \theta} | \mathcal{D}) = \delta({\bm \theta} - \hat{\bm \theta}) \implies 
p({\bm \mu}| {\bm x}^*, \mathcal{D}) \approx p({\bm \mu}| {\bm x}^*, \hat{\bm \theta})$.
\normalsize

\textbf{Construction. }
A Dirichlet distribution is parameterized using a vector, called the concentration parameters, \small{$\bm{\alpha}= {\{\alpha_1, \cdots, \alpha_K\}}$}, 
\normalsize{as:}
\small{
$Dir({\bm \mu | \bm \alpha}) = \frac{\Gamma(\alpha_0)}{\prod_{c=1}^{K}\Gamma(\alpha_c)}\prod_{c=1}^{K} \mu_c^{\alpha_c-1}, ~~ 
\alpha_c >0 $,}
\normalsize{where,} \small{$\alpha_0 = \sum_{c=1}^K \alpha_c$) 
\normalsize{denotes its precision. }
A larger precision leads to a sharper Dirichlet distribution. 

A DPN, $f_{\bm \hat{\theta}}$ produces the concentration parameters, 
$\bm{\alpha}= {\{\alpha_1, \cdots, \alpha_K\}}$ corresponding to each class, i.e, 
$\bm{\alpha} = f_{\bm \hat{\theta}}({\bm x^*})$.
The posterior over class labels is given by the mean of the Dirichlet, i.e,
\begin{equation} \small
\label{eq:dpn_construct}
p(\omega_c | {\bm x^*}; \hat{\bm \theta})
= \int p(\omega_c|{\bm \mu})~~ p({\bm \mu| \bm x^*; \hat{\bm \theta} }) ~~ d{\bm \mu} ~~
= \frac{\alpha_c}{\alpha_0},
\qquad \text{where, ~~} p({\bm \mu|\bm x^*; \bm \hat{\theta} }) = Dir({\bm \mu | \bm \alpha})
\end{equation}
A standard DNN with the softmax activation function can be represented as a DPN where the concentration parameters are $\alpha_c = e^{z_c({\bm x^*})}$;
$z_c(\bm x^*)$ is the pre-softmax (logit) output corresponding to the class, $c$ for an input ${\bm x^*}$.
The expected posterior probability of class label $\omega_c$ is given as:
\begin{equation}\small
\label{eq:softmax_dpn}
p(\omega_c | {\bm x^*}; \hat{\bm \theta}) ~~
= \frac{\alpha_c}{\alpha_0}
= \frac{e^{z_c({\bm x^*})}}{\sum_{c=1}^{K} e^{z_c({\bm x^*})}}
\end{equation}
However, the mean of the Dirichlet distribution is now \textit{insensitive} to any arbitrary scaling of $\alpha_c$.
Hence, while the standard \textit{cross-entropy loss} efficiently models the mean of the Dirichlet distributions, it degrades the precision, $\alpha_0$.

Malinin \& Gales (2018)  \cite{dpn_nips_2018} propose a forward KL (FKL) divergence loss that explicitly minimizes the KL divergence between the model and the given target Dirichlet distribution.
Malinin \& Gales (2019) \cite{dpn2_nips_2019} further propose a reverse KL (RKL) loss function that reverses the terms in the KL divergence to induce a uni-modal Dirichlet as the target distribution and improve their scalability for classification tasks with a larger number of classes.
The RKL loss trains a DPN using both in-domain and OOD training examples in a multi-task fashion:
\begin{equation} \small \displaystyle
\label{eq:dpn_2019}
\mathcal{L}^{rkl}(\theta; \gamma, 
{\bm\beta}^y, {\bm\beta}^{out})
=
\mathbb{E}_{P_{in}} \text{KL}[p({\bm \mu}| {\bm x}, {\bm \theta}) || Dir({\bm \mu} | {\bm \beta}^y)] 
+ \gamma \cdot \mathbb{E}_{P_{out}}\text{KL}[p({\bm \mu}| {\bm x}, {\bm \theta}) || Dir({\bm \mu} | {\bm \beta}^{out}) ]
\end{equation}
where $P_{in}$ and $P_{out}$ are the distribution for the in-domain and OOD training examples and 
${\bm \beta}^{y}$ and ${\bm \beta}^{out}$ their hand-crafted target concentration parameters respectively.
\section{Proposed Methodology}
\label{sec:proposed}
\textbf{Shortcomings of DPN using RKL loss.}
We first demonstrate that the RKL loss function tends to produce flatter Dirichlet distributions for in-domain misclassified examples, compared to its confident predictions. 
We can decompose the reverse KL-divergence loss into two terms i.e
\textit{reverse cross entropy}, 
\small
$\displaystyle \mathbb{E}_{P({\bm \mu}| {\bm x}, {\bm \theta})} [-\ln Dir ({\bm \mu} | \bm{\overline{\beta}})]$ 
\normalsize
and \textit{differential entropy}, 
\small
$\displaystyle \mathcal{H}[p({\bm \mu}| {\bm x}, {\bm \theta})]$, 
\normalsize
as shown in \cite{dpn2_nips_2019}:
\begin{equation} \small
\label{eq_rev_kl}
\displaystyle 
\mathbb{E}_{ \tilde{P}_{T}(\bm x,y)} ~ \text{KL} \big[ ~ p({\bm \mu}| {\bm x}, {\bm \theta}) ~||~ Dir({\bm \mu} | {\bm \beta}) ~ \big] 
= \mathbb{E}_{ \tilde{P}_{T}(\bm x)} \Big[ \mathbb{E}_{P({\bm \mu}| {\bm x}, {\bm \theta})} [-\ln Dir ({\bm \mu} | \bm{\overline{\beta}})] - \mathcal{H} \big[p({\bm \mu}| {\bm x}, {\bm \theta}) \big] \Big]
\end{equation}
\small
where $\bm{\beta} = \{\beta^{(c)}_1, \cdots, \beta^{(c)}_K \}$ 
\normalsize
represents their hand-crafted target concentration parameters, and
$\bm{\overline{\beta}}$ represents the concentration parameter of the expected target Dirichlet with respect to the empirical training distribution $\tilde{P}_{T}$. 
In our analysis, we replace $\tilde{P}_{T}$ with the empirical distribution of in-domain training examples $\tilde{P}_{in}$ or OOD training examples $\tilde{P}_{out}$.

Differential entropy measures the sharpness of a continuous distribution.
Minimizing $\displaystyle - \mathcal{H}[p({\bm \mu}| {\bm x}, {\bm \theta})]$ always leads to produce a flatter distribution. 
Hence, we rely only on $\displaystyle \mathbb{E}_{P({\bm \mu}| {\bm x}, {\bm \theta})} [-\ln Dir ({\bm \mu} | \bm{\overline{\beta}})]$ to produce sharper distributions.
Malinin \& Gales (2019) \cite{dpn2_nips_2019} choose the target concentration value for in-domain examples as: $(\beta + 1)$ for the correct class and $1$ for the incorrect classes.
Thus, we get:
\begin{equation} \small
\label{eq_rev_kl_2}
\begin{split}
\mathbb{E}_{ \tilde{P}_{T}(\bm x)} \Big[ \mathbb{E}_{P({\bm \mu}| {\bm x}, {\bm \theta})} [-\ln Dir ({\bm \mu} | \bm{\overline{\beta}})] \Big]
& 
= \mathbb{E}_{\tilde{P}_{T}(\bm x)}\Big[\sum_{c} \sum_{k} \tilde{p}(\omega_c|\bm{x}) (\beta^{(c)}_k - 1)\big[ \psi(\alpha_0)-\psi(\alpha_k) \big]\Big] \\
& 
= \mathbb{E}_{\tilde{P}_{T}(\bm x)} \Big[ \beta \psi(\alpha_0)
-\sum_{c} \beta \tilde{p}(\omega_c|\bm{x}) \psi(\alpha_c)\Big]
\end{split}
\end{equation}
where $\psi$ is the digamma function. 

We can see in Eqn. \ref{eq_rev_kl_2}, the reverse cross-entropy term maximizes $\psi(\alpha_c)$ for each class $c$ with the factor, $\beta \tilde{p}(\omega_c|\bm{x})$, and minimizes $\psi(\alpha_0)$ with the factor, $\beta$.
For an in-domain example with confident prediction, it produces a sharp Dirichlet with a large concentration value for the correct class and very small concentration parameters ($<<1$) for the incorrect classes.
However, for an input with high data uncertainty, $\beta$ is distributed among multiple classes according to $\tilde{p}(\omega_c|\bm{x})$. 
This leads to relatively smaller (but $\geq 1$) concentration parameters for all overlapping classes, producing a much flatter and diverse Dirichlet.

\begin{wrapfigure}{r}{7.5cm}
\vspace*{-1.0em}
\caption{Dirichlet distributions with the same precision but different concentration parameters.}
\centering
\begin{subfigure}[t]{0.54\linewidth}
\centering
\captionsetup{justification=centering}
\includegraphics[width=0.5\linewidth, height=40pt]{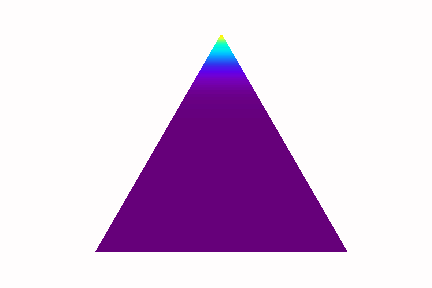}
\centering
\caption*{\small(a) $Dir_1$:$\{0.01,0.01,101.98\}$
		\\ \small D.Ent = -199.1, MI=$8e$-4}
\end{subfigure} 
\begin{subfigure}[t]{0.44\linewidth}
\centering
\captionsetup{justification=centering}
\includegraphics[width=0.7\linewidth, height=40pt]{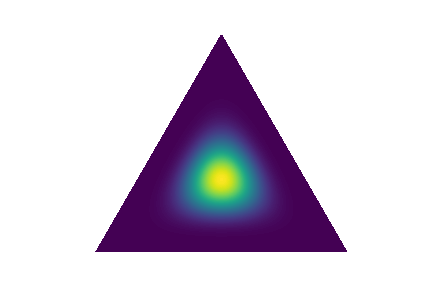}
\centering
\caption*{\small(b) $Dir_2$:$\{34, 34, 34\}$
		\\ \small D.Ent = -3.45, MI=$9.7e$-3}
\end{subfigure}
\vspace*{-1.0em}
\label{fig:same_precision}
\end{wrapfigure}
For example, let us consider two Dirichlet distributions, $Dir_1$ and $Dir_2$, with the same precision, $\alpha_0=102$, but different concentration parameters of $\{0.01,0.01,101.98\}$ (Fig. \ref{fig:same_precision}a) and $\{34, 34, 34\}$ (Fig. \ref{fig:same_precision}b).
We measure the differential entropy (D.Ent) and mutual information (MI), which are maximized for flatter and diverse distributions respectively.
We can see that $Dir_1$ produces much lower D.Ent and MI scores than $Dir_2$.
It shows that $Dir_2$ is flatter and diverse than $Dir_1$, even with the same precision values.
The differences in these scores become more significant in higher dimensions.
As we consider the same example for a classification task with $K=100$, D.Ent and MI would respectively produce $-9.9e$3 and $0.02$ for $Dir_1$ and $-370.5$ and $0.23$ for $Dir_2$.
Further, in Section \ref{sec:proposed} (and in Table \ref{table:misclassification} (Appendix)), we show that DPN models also tend to produce lower precision values along with flatter and diverse Dirichlet distributions for misclassified examples.

This behavior is \textit{not} desirable:
since the RKL loss also trains the DPN to produce flatter and diverse Dirichlet distributions for OOD examples, it often leads to indistinguishable distributions for OOD examples in the boundary cases and in-domain misclassified examples.
However, we can produce sharper Dirichlet distributions for OOD examples, as in Figure \ref{fig:dirichlet_2(d)}, to maximize their representation gap from in-domain examples.
For OOD training examples, we should choose identical values for target concentration parameters, $(\tau+1)$ where $\tau > -1$, for all classes to produce uniform categorical posterior distributions.
Using $(\tau+1)$ for $\beta^{(c)}_k$ in Eqn. \ref{eq_rev_kl_2}, we get the RKL loss for OOD examples as:
\begin{equation} \small
\label{eq:rkl_ood}
\begin{split}
\mathbb{E}_{P_{T}(\bm x)}\Big[ \tau K \psi(\alpha_0) - \sum_{c} \tau \psi(\alpha_c) - \mathcal{H} \big[p({\bm \mu}| {\bm x}, {\bm \theta}) \big] \Big]  
\end{split}
\end{equation}
Malinin \& Gales (2019) \cite{dpn2_nips_2019} choose  $\tau$ to $0$. 
This results the RKL loss to minimize $- \mathcal{H}[p({\bm \mu}| {\bm x}, {\bm \theta})]$.
Hence, the DPN produces flat Dirichlet distributions for OOD examples.
In the following, we investigate the other choices of $\tau$.

Choosing $\tau > 0$ gives an objective function that minimizes the precision, \small{$\alpha_0 (= \sum_{c=1}^{K} \alpha_c)$} 
\normalsize
of the output Dirichlet distribution while maximizing individual concentration parameters, $\alpha_c$ (Eq. \ref{eq:rkl_ood}).
In contrast, choosing $\tau \in (-1,0)$ maximizes  $\alpha_0$ while minimizing $\alpha_c$'s.
Hence, we can conclude that either choice of $\tau$ may lead to \textit{uncontrolled values} for the concentration parameters for an OOD example.
We also empirically verify this analysis in Appendix \ref{sec:rkl_fraction}.


\textbf{Explicit Precision Regularization. }
We propose a new loss function for DPN classifiers that separately models the mean and precision of the output Dirichlet to achieve greater control over the desired behavior of a DPN classifier. 
We  model the mean of the output Dirichlet using soft-max activation function and the cross-entropy loss, as in Eqn. \ref{eq:softmax_dpn}, along with a novel \textit{explicit precision regularizer}, that controls the precision.

\begin{wrapfigure}{r}{6.3cm}
\centering
\vspace{-1.0em}
\caption{Growth of regularizers w.r.t logits.}
\label{fig:regularizers}
\includegraphics[width=\linewidth, height=65pt]{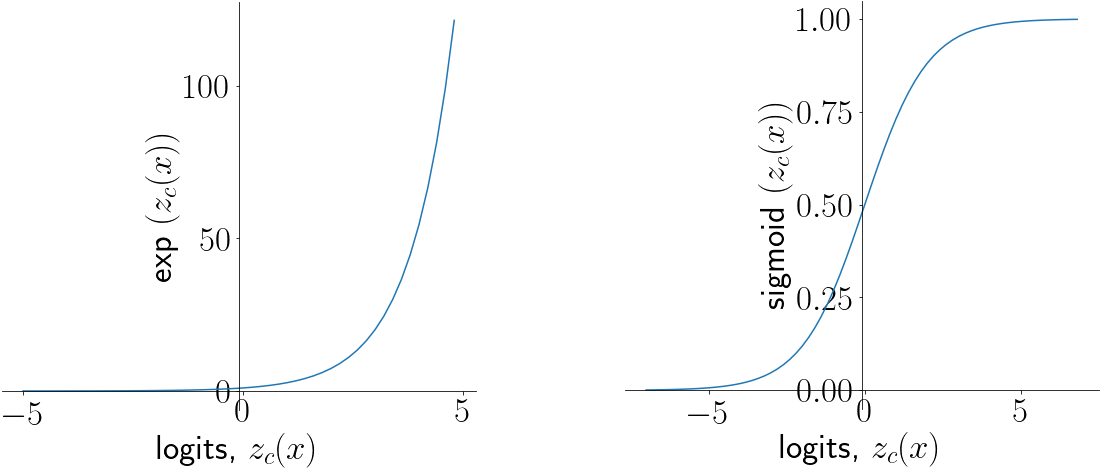}
\caption*{\small (a) $\sum_c \exp{z_c(\bm x)}$ \hspace*{1.5em}
(b) $\frac{1}{K}\sum_c \text{sigmoid}{z_c(\bm x)}$ }
\vspace{-1.0em}
\end{wrapfigure}

Note that, we cannot choose the regularizer that directly maximizes the precision, \small{$\alpha_0 = \sum_{c=1}^K \exp{z_c({\bm x})}$ } \normalsize to control the output Dirichlet distributions.
This is because the term, $\exp{z_c({\bm x})}$ is unbounded.
Hence, using the precision, $\alpha_0 =\sum_c \exp{z_c(\bm x)}$ as the regularizer leads to large logit values for in-domain examples (see Figure \ref{fig:regularizers}(a)). 
However, it would make the cross-entropy loss term negligible, degrading the  in-domain classification accuracy.
Further, $\exp{z_c(\bm x)}$ is not a symmetric function. 
Hence, it does not equally constrain the network to produce small fractional concentration parameters, i.e $\alpha_c = \exp{z_c(\bm x)} \rightarrow 0$, for OOD examples to produce the desirable multi-modal Dirichlet distributions (Figure \ref{fig:dirichlet_2(d)}).
Moreover, in practice, the choice of $\sum_c \exp{z_c(\bm x)}$ leads the training loss to NaN.

In contrast, by limiting logits, $z_c$ to values that are, for example, approximately $5$ for in-domain examples, and $-5$  for OOD examples, we would have the desirable sharp uni-modal and multi-modal Dirichlet distributions respectively, maximizing their representation gaps (see Figure \ref{fig:dirichlet_2}). 
Beyond these values, the cross-entropy loss should become the dominant term in the loss function to improved the in-domain classification accuracy. 

Hence, we propose a logistic-sigmoid approximation to \textit{individually control} the concentration parameters using $\frac{1}{K}\sum_{c=1}^K\text{sigmoid}(z_c({\bm x}))$ as the regularizer to control the spread of the output Dirichlet distributions.
This regularizer is applied alongside the cross-entropy loss on the soft-max outputs.
The use of logistic-sigmoid function satisfies this condition by providing an implicit upper and lower bounds on the individual concentration parameters for both in-domain and OOD examples (see Figure \ref{fig:regularizers}(b)).
For interval $(-\infty, \epsilon)$,  when $\epsilon$ is close to $0$, the approximation error is low.
While for the interval $[\epsilon, \infty )$, it offers the desired behavior of monotonically increasing function, similar to the exponential function, however within a finite boundary.

The proposed loss function for the in-domain training examples is given as:
\begin{equation}
\label{eq:l_in_reg}
\small
\mathcal{L}_{in}({\bm \theta}, \lambda_{in}) :=
\mathbb{E}_{P_{in}(\bm x, y)} \Big[- \log p({y} | {\bm x}, {\bm \theta})
- \frac{\lambda_{in}}{K}\sum_{c=1}^K\text{sigmoid}(z_c({\bm x})) \Big]
\end{equation}
Similarly, for OOD training examples, we can control precision values using the loss function as:
\begin{equation}
\label{eq:l_out_reg}
\small
\mathcal{L}_{out}({\bm \theta}, \lambda_{out}) :=
\mathbb{E}_{P_{out}(\bm x, y)} \Big[\mathcal{H}_{ce} ( \mathcal{U}; p( {y} | {\bm x}, {\bm \theta}))
-  \frac{\lambda_{out}}{K}\sum_{c=1}^K\text{sigmoid}(z_c({\bm x})) \Big]
\end{equation}
where $\mathcal{H}_{ce}$ is the cross-entropy function.
$\mathcal{U}$ is the uniform distribution over the class labels. 
$\lambda_{in}$ and $\lambda_{out}$ are user-defined hyper-parameters for the regularization terms to control the precision of the output distributions.
We train the DPN in a multi-task fashion using the overall loss function as:
\begin{equation}
\label{eq:l_total_reg}
\small
\min_{\bm \theta} \mathcal{L}({\bm \theta}; \gamma, \lambda_{in}, \lambda_{out}) = \mathcal{L}_{in}({\bm \theta}, \lambda_{in}) 
+ \gamma \mathcal{L}_{out}({\bm \theta}, \lambda_{out})
\end{equation}
%
where $\gamma >0$ balances between the loss values for in-domain examples and OOD examples.
We now analyze the proposed regularizer by taking expectation with respect to the empirical distribution, $\tilde{P}_T$:
\begin{equation} \small
\begin{split} 
\mathbb{E}_{\tilde{P}_{T}(\bm x, y)} \Big[- \frac{\lambda_{T}}{K}\sum_{c=1}^K\text{sigmoid}(z_c({\bm x})) \Big] 
& = \mathbb{E}_{P_{T}(\bm x)} \Big[- \frac{\lambda_{T}}{K} \sum_{k=1}^K p(y=\omega_k|\bm x) \Big[\sum_{c=1}^K  \text{sigmoid}(z_c({\bm x})) \Big] \Big] \\
& = \mathbb{E}_{P_{T}(\bm x)} \Big[- \frac{\lambda_{T}}{K}\sum_{c=1}^K \text{sigmoid}(z_c({\bm x})) \Big]
\end{split}
\end{equation}
where $\lambda_T$ and $\tilde{P}_{T}$ should be replaced to $\lambda_{in}$ and $\tilde{P}_{in}$ respectively for in-domain examples, and $\lambda_{out}$ and $\tilde{P}_{out}$ respectively for OOD examples.

Hence, by choosing $\lambda_{in}>0$ for in-domain examples, our regularizer imposes the network to maximize $\text{sigmoid}(z_c({\bm x}))$ irrespective of the class-labels.
However, for confidently predicted examples, the cross-entropy loss ensures to maximize the logit value for the correct class.
In contrast, in the presence of high data uncertainty, the cross-entropy loss produces a multi-modal categorical distribution over the overlapping classes.
Hence, as before, it leads to producing a flatter distribution for misclassified examples.
Now, by choosing $\lambda_{in} > \lambda_{out}>0$, we also enforce the network to produce a flatter distribution with $\alpha_c = \exp{z_c(\bm x^*)} \geq 1$ for an OOD example $\bm x^*$.
Hence,  the DPN will produce indistinguishable representations for an in-domain example with high data uncertainty as an OOD example, similar to the RKL loss (Eq. \ref{eq_rev_kl}).

\begin{wrapfigure}{r}{8.cm}
\centering
\vspace{-1.0em}
\caption{Dirichlet distributions with different precision.}
\begin{subfigure}[t]{0.3\linewidth}
\centering
\captionsetup{justification=centering}
\includegraphics[width=0.9\linewidth, height=40pt]{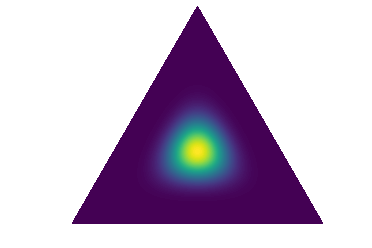}
\caption*{(a) \small{${\bm \alpha}$= \{10, 10, 10\} }
		\\ \small D.Ent=-2.3 
		\\ \small MI=0.03}
\end{subfigure}
\begin{subfigure}[t]{0.3\linewidth}
\centering
\captionsetup{justification=centering}
\includegraphics[width=0.9\linewidth, height=40pt]{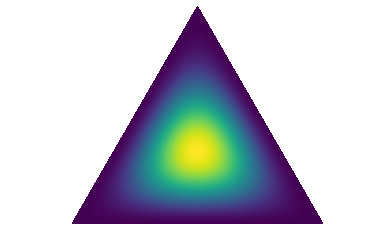}
\caption*{(b) \small{${\bm \alpha}$= (3, 3, 3) } 
		\\ \small D.Ent=-1.2 
		\\ \small MI=0.1}
\end{subfigure}
\begin{subfigure}[t]{0.33\linewidth}
\centering
\captionsetup{justification=centering}
\includegraphics[width=0.9\linewidth, height=40pt]{./images/dir_fraction.png}
\caption*{(c) \small ${\bm \alpha}$= (0.1, 0.1, 0.1)  
		\\ \small D.Ent=-13.0, 
		\\ \small MI=0.84}
\end{subfigure}
\vspace{-1.0em}
\label{fig:dirichlet_3}
\end{wrapfigure}
However, now we can address this problem by choosing $\lambda_{out} < 0$. 
It enforces the DPN to produce negative values for $z_c(\bm x^*)$ and thus \textit{fractional} values for $\alpha_c$'s for OOD examples.
This leads the probability densities to be moved across the edges of the simplex to produce extremely sharp multi-modal distributions, solving the original problem described in the beginning.

For example, let a DPN with $\lambda_{in},\lambda_{out}> 0$ represents a misclassified example (Fig. \ref{fig:dirichlet_3}a) and an OOD example (Fig. \ref{fig:dirichlet_3}b).
We can see that their representations are very similar, even if their concentration parameters are different.
In contrast, our DPN with $\lambda_{in}>0, \lambda_{out} < 0$ leads to a sharp, multi-modal Dirichlet as in Fig. \ref{fig:dirichlet_3}(c) for an OOD, maximizing the \textit{representation gap} from Fig. \ref{fig:dirichlet_3}(a).
We can confirm this by observing their \textit{mutual information (MI)} scores.
However, the choice of $\lambda_{in} = 0$ and $\lambda_{out}<0$ does not enforce these properties (see  ablation study in Appendix \ref{sec:lambda_0}).

The overall loss function in Eqn. \ref{eq:l_total_reg} requires training samples from both in-domain distribution and OOD.
Here, we select a different real-world dataset as our OOD training examples. 
It is more feasible in practice and performs better than the artificially generated OOD examples \cite{oe_iclr_2019,gan_ood_2018}.
\section{Experimental Study}
We conduct two sets of experiments: 
First, we experiment on a synthetic dataset.
Next, we present a comparative study on a range of image classification tasks.
\footnote{\small Please refer to Appendix for additional results and ablation studies.}
\footnote{\small Code Link: \href{https://github.com/jayjaynandy/maximize-representation-gap}
{\textsf{https://github.com/jayjaynandy/maximize-representation-gap}}}
\subsection{Synthetic Dataset}
\label{sec:toy_main}
We construct a simple synthetic dataset with three overlapping classes to study the characteristics of different DPN models.
We sample the in-domain training instances from three different overlapping isotropic Gaussian distributions, to obtain these overlapping classes, as shown in Figure \ref{fig:toy_experiment}a.
We demonstrate the results of our DPN models with $\lambda_{in} > 0$ and both positive and negative values for $\lambda_{out}$, are denoted as DPN$^+$ and DPN$^-$ respectively.
See Appendix \ref{sec:toy_app} for additional details on experimental setup, hyper-parameters and the results using RKL loss function \cite{dpn2_nips_2019}.

A \textit{total predictive uncertainty} measure is derived from the expected predictive categorical distribution, $p(\omega_c| \bm{x}^*, D)$ i.e by marginalizing $\bm \mu$ and $\bm \theta$ in Eq. \ref{eq:dpn}.
Maximum probability in the expected predictive categorical distribution is a popular total uncertainty measure: 
$max\mathcal{P} =  \max_c p(\omega_c| \bm{x}^*, D)$.

\begin{figure}
\vspace{-1.3em}
\caption{ Uncertainty measures of different data-points for DPN$^+$ (top row) and DPN$^-$ (bottom row).}
\label{fig:toy_experiment}
\begin{minipage}{.2\textwidth}
\begin{subfigure}{\linewidth}
\centering
\captionsetup{justification=centering}
\caption*{ \small (a) In-domain \\training data}
\label{fig:toy_data}
\includegraphics[width=1\linewidth,height=75pt]{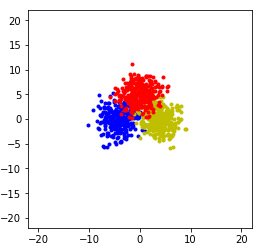}
\end{subfigure}
\end{minipage}
\begin{minipage}{.75\textwidth}
\begin{subfigure}{\linewidth}
\centering
\includegraphics[width=1\linewidth,height=65pt]{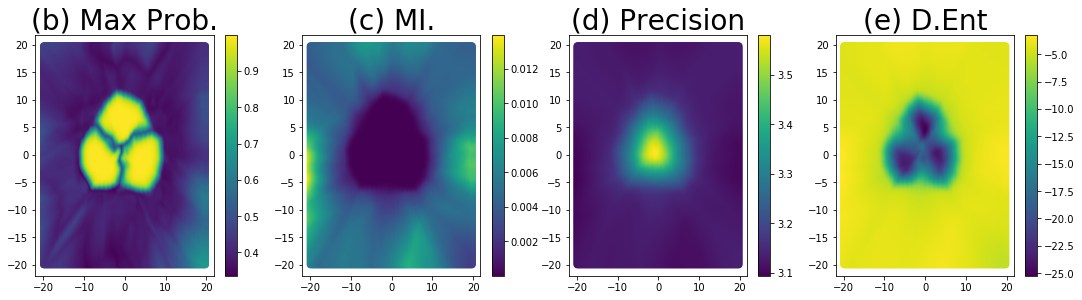}
\label{fig:toy_d_ent}
\vspace{-0.9em}
\end{subfigure}\\
\begin{subfigure}{\linewidth}
\centering
\includegraphics[width=1\linewidth,height=65pt]{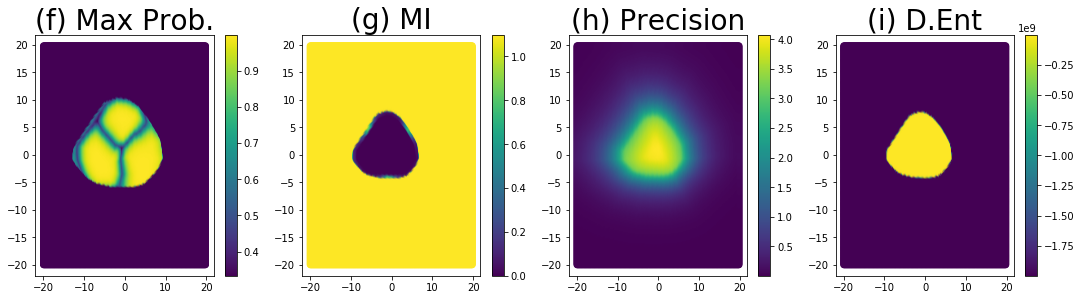}
\label{fig:toy_d_ent_2}
\end{subfigure}
\end{minipage}%
\vspace{-1.3em}
\end{figure}
Fig \ref{fig:toy_experiment}b and Fig \ref{fig:toy_experiment}f show the Max.P uncertainty measures for different data points for DPN$^+$ and DPN$^-$ respectively.
We can see that DPN$^-$ appropriately interpolates the concentration parameters to maximize the margin on the boundary of the in-domain and OOD regions, leading to improved  OOD detection performance even using total uncertainty measures.
However, since the predicted distributions are obtained by marginalizing ${\bm \mu}$ (Eqn. \ref{eq:dpn}), the total uncertainty measures fail to robustly distinguish the OOD examples from the misclassified examples.
As we can see in both Fig \ref{fig:toy_experiment}b and Fig \ref{fig:toy_experiment}f that the Max.P produces lower scores in the class overlapping regions.
Since the non-Bayesian models only relies on the total uncertainty measures, they are unable   to reliably distinguish between data and distributional uncertainty \cite{dpn_nips_2018}.

A DPN can address this limitation by computing the \textit{mutual information (MI)} between $y$ and $\bm \mu$ i.e 
\small{$\mathcal{I}[y, \bm \mu | \bm{x^*}, \bm{\hat \theta}]$}.
\normalsize{}
We can also measure the \textit{expected pairwise KL divergence (EPKL)} between the pairs of independent ``categorical distribution" samples from the Dirichlet \cite{malinin_thesis_2019}. 
For a Dirichlet distribution, EPKL is simplified to $\frac{K-1}{\alpha_0}$, where $\alpha_0$ is the precision \cite{malinin_thesis_2019}.
Since a DPN also produces smaller precision values for OOD examples, we can directly view the \textit{precision} as a distributional uncertainty measure.
Note that, both EPKL and \textit{precision (or inverse-EPKL)} leads to the same OOD detection performance as they produce the same relative uncertainty scores (in reverse order) for a given set of test examples.
In Fig \ref{fig:toy_experiment}c and Fig \ref{fig:toy_experiment}d, we observe that our DPN$^+$ model successfully distinguishes the OOD examples using the mutual information and the precision measures respectively.
However, our DPN$^-$ model clearly demonstrates its superiority by producing sharper and significant differences of uncertainty scores for OOD examples, compared to the in-domain examples (Fig \ref{fig:toy_experiment}g and Fig \ref{fig:toy_experiment}h).

\textit{Differential entropy (D.Ent)}, $\mathcal{H}[p({\bm \mu}| \bm x^*, \bm \hat \theta)]$, that maximizes for sharp Dirichlet distributions, is also used as a distributional uncertainty measure \cite{dpn_nips_2018,malinin_thesis_2019}. 
However, unlike other DPN models, our DPN$^-$ behaves differently to produce sharp multi-modal Dirichlet distributions for OOD examples. 
Hence, D.Ent also behaves in an \textit{inverted manner}, compared to the other DPN models.
As we can see in Fig \ref{fig:toy_experiment}e, DPN$^+$ produces higher D.Ent values for OOD examples.
In contrast, DPN$^-$ produces \textit{large negative} D.Ent scores for OOD examples, indicating that it often produces even sharper Dirichlet for OOD examples, than the confidently predicted examples  (Fig \ref{fig:toy_experiment}i).

\subsection{Benchmark Image Classification Datasets}
\label{sec:benchmark}
Next, we carry out experiments on  CIFAR-10 and CIFAR-100 \cite{db_cifar} and TinyImageNet \cite{db_tiny}.
We train the C10 classifiers by using CIFAR-10 training images as in-domain data and CIFAR-100 training images as OOD data.
C100 classifiers are trained by using CIFAR-100 training images as in-domain and CIFAR-10 training images as OOD.
For the TIM classifier, we use the TinyImageNet images as in-domain training data and ImageNet-25K images as OOD training data.
ImageNet-25K is obtained by randomly selecting $25,000$ images from the ImageNet dataset \cite{db_imagenet}.
We use the VGG-16 network for these tasks \cite{vgg16}.
Here, we study the performance our DPN models with $\lambda_{in} >0$ and both $\lambda_{out} > 0$ and $\lambda_{out} < 0$, are denoted as DPN$^+$ and DPN$^-$ respectively. 
See Appendix~\ref{sec:ablation} for additional ablation studies.
We compare our models with the standard DNN \cite{baseline_iclr_2017}, Bayesian MCDP  \cite{mcDrop_icml_gal16}, Deep Ensemble (DE) \cite{ensemble_nips_2017},
non-Bayesian OE \cite{oe_iclr_2019} and the existing DPN$_{rev}$ model \cite{dpn2_nips_2019}.
For Bayesian MCDP and DE, we can compute the mutual information (MI). 
However, we cannot compute the precision or D.Ent for them.
For the non-Bayesian models, MI, precision, and D.Ent are not defined.
See Appendix \ref{sec:benchmark_app} for our experimental details along with additional discussions.

We present the performances of our models for \textit{OOD detection} and \textit{ misclassification detection} tasks.  
Note that the in-domain and OOD test examples are kept separately from the training examples, as in a real-world scenario (see Table \ref{table:details_1}(Appendix)).
For OOD detection, we choose the OOD examples as the `positive' class and in-domain examples as the `negative' class. 
For misclassification detection, we consider the misclassified examples as the `positive' class and correctly classified examples as the `negative' class. 
Here, we use \textit{area under the receiver operating characteristic (AUROC)} metric \cite{baseline_iclr_2017}.
We present the results using Max.P, MI, $\alpha_0$ (or inverse-EPKL), and D.Ent.
We report the (mean $\pm$ standard deviation) of three different models.
We provide additional results including classification accuracy, and performance on a wide range of OOD datasets along with area under the precision-recall curve (AUPR) metric and entropy measure in Appendix \ref{sec:extended} (Table \ref{table:misclassification}-\ref{table:extended_result_tim}).

\begin{table}[h]	
	\centering
	\caption{AUROC scores for OOD detection (mean $\pm$ standard deviation of 3 runs).
	Refer to Table \ref{table:extended_result_cifar10}-\ref{table:extended_result_tim}
	(Appendix) for AUPR scores and results on additional OOD test sets.}
\label{table:ood}
	\resizebox{14.0cm}{!}{%
		\begin{tabular}{ll|cccc|cccc|cccc}
			\hline
			& \multicolumn{1}{c|}{OOD} 
			& \multicolumn{4}{c}{Tiny \cite{db_tiny} }
			& \multicolumn{4}{c}{STL-10 \cite{db_stl10} }
			& \multicolumn{4}{c}{LSUN \cite{db_lsun} }  \\
			
			& \multicolumn{1}{c|}{} 
			& Max.P & MI & $\alpha_0$ & D.Ent
			& Max.P & MI & $\alpha_0$ & D.Ent
			& Max.P & MI & $\alpha_0$ & D.Ent\\
			\cline{2-14}
			& Baseline
			& 88.9{\tiny$\pm$0.0} & -  & -  & -  
			& 75.9{\tiny$\pm$0.0} & - & -  & - 
			& 90.3{\tiny$\pm$0.0} & -  & -  & - 
			\\
			
			\multirow{6}{*}{\rotatebox{90}{C10}}
			& MCDP 
			& 88.7{\tiny$\pm$0.1} & 88.1{\tiny$\pm$0.1}  &- &- 
			& 76.2{\tiny$\pm$0.0} & 76.0{\tiny$\pm$0.0} &- &- 
			& 90.6{\tiny$\pm$0.0} & 90.2{\tiny$\pm$0.0} &- &- 
			\\
			
			& DE 
			& 88.9{\tiny$\pm$NA} & 87.8{\tiny$\pm$NA} &- &- 
			& 76.0{\tiny$\pm$NA} & 75.6{\tiny$\pm$NA} & - &- 
			& 90.3{\tiny$\pm$NA} & 89.7{\tiny$\pm$NA} &- &- 
			\\
			

			& OE 
			& 98.2{\tiny$\pm$0.1} & -  & - & -  
			& 81.4{\tiny$\pm$1.2} & -  & -  & -  
			& 98.4{\tiny$\pm$0.3} & -  & -  & -  
			\\
			
			& DPN$_{rev}$ 
			& 97.5{\tiny$\pm$0.5}  & 97.8{\tiny$\pm$0.4}  & 97.8{\tiny$\pm$0.4}  & 97.7{\tiny$\pm$0.4}  
			& 81.6{\tiny$\pm$1.7}  & 82.2{\tiny$\pm$1.7}  & 82.2{\tiny$\pm$1.6}  & 81.9{\tiny$\pm$1.7}  
			& 98.5{\tiny$\pm$0.4}  & 98.7{\tiny$\pm$0.3}  & 98.7{\tiny$\pm$0.3}  & 98.7{\tiny$\pm$0.3}  
			\\
			
			& DPN$^+$ 
			& 98.0{\tiny$\pm$0.2} & 98.0{\tiny$\pm$0.2}  & 98.0{\tiny$\pm$0.2}  & 98.0{\tiny$\pm$0.2}  
			& 81.6{\tiny$\pm$1.4}  & 81.8{\tiny$\pm$1.2}  & 81.8{\tiny$\pm$1.2}  & 81.8{\tiny$\pm$1.2}  
			& 98.2{\tiny$\pm$0.3}  & 98.3{\tiny$\pm$0.4}  & 98.3{\tiny$\pm$0.4}  & 98.3{\tiny$\pm$0.4}  
			\\
			
			& DPN$^-$ 
			& \textbf{99.0{\tiny$\pm$0.1}}  & \textbf{99.0{\tiny$\pm$0.1} } & 97.7{\tiny$\pm$0.1}  & 6.0{\tiny$\pm$0.3}  
			& 84.7{\tiny$\pm$0.4}  & \textbf{85.3{\tiny$\pm$0.5}}  & 84.9{\tiny$\pm$0.5}  & 34.6{\tiny$\pm$0.4}  
			& 99.2{\tiny$\pm$0.1}  & \textbf{99.3{\tiny$\pm$0.0}}  & 98.1{\tiny$\pm$0.1}  & 5.0{\tiny$\pm$0.2}  
			\\ \hline
		\end{tabular}
	}
	\resizebox{14.0cm}{!}{%
		\begin{tabular}{ll|cccc|cccc|cccc}
			\hline
			& \multicolumn{1}{c|}{OOD} 
			& \multicolumn{4}{c}{Tiny \cite{db_tiny}}
			& \multicolumn{4}{c}{STL-10 \cite{db_stl10} }
			& \multicolumn{4}{c}{LSUN \cite{db_lsun} }  \\

			& \multicolumn{1}{c|}{} 
			& Max.P & MI & $\alpha_0$ & D.Ent
			& Max.P & MI & $\alpha_0$ & D.Ent
			& Max.P & MI & $\alpha_0$ & D.Ent\\
			\cline{2-14}
			
			& Baseline
			& 68.8{\tiny$\pm$0.2} & -  & -  & - 
			& 69.6{\tiny$\pm$0.0} & - & - & - 
			& 72.5{\tiny$\pm$0.0} & -  & - & -
			\\
			
			\multirow{6}{*}{\rotatebox{90}{C100} } 
			& MCDP 
			& 69.7{\tiny$\pm$0.3}  & 70.6{\tiny$\pm$0.3} &- &- 
			& 70.7{\tiny$\pm$0.1}  & 71.6{\tiny$\pm$0.2} &- &- 
			& 74.5{\tiny$\pm$0.1}  & 75.9{\tiny$\pm$0.2} &- &- 
			\\
			
			& DE 
			& 68.9{\tiny$\pm$NA} & 69.6{\tiny$\pm$NA} &- &- 
			& 69.6{\tiny$\pm$NA} & 70.2{\tiny$\pm$NA} &- &-
			& 72.6{\tiny$\pm$NA} & 73.4{\tiny$\pm$NA} &- &- 
			\\
			

			& OE 
			& 89.5{\tiny$\pm$1.0}  & - & - & - 
			& 91.2{\tiny$\pm$0.7}  & - & - & - 
			& 92.2{\tiny$\pm$0.9}  & - & - & - 
			\\
			
			& DPN$_{rev}$ 
			& 81.2{\tiny$\pm$0.2}  & 83.8{\tiny$\pm$0.1}  & 83.8{\tiny$\pm$0.1}  & 83.5{\tiny$\pm$0.1}  
			& 87.2{\tiny$\pm$0.1}  & 89.3{\tiny$\pm$0.1}  & 89.3{\tiny$\pm$0.1}  & 89.0{\tiny$\pm$0.1}   
			& 86.7{\tiny$\pm$0.0} & 89.3{\tiny$\pm$0.1}  & 89.3{\tiny$\pm$0.1}  & 88.9{\tiny$\pm$0.1}  
			\\
			
			& DPN$^+$ 
			& 85.9{\tiny$\pm$0.3}  & 92.2{\tiny$\pm$0.1}  & 92.2{\tiny$\pm$0.1}  & 92.3{\tiny$\pm$0.1}  
			& 89.1{\tiny$\pm$0.2}  & 95.0{\tiny$\pm$0.0}  & 95.0{\tiny$\pm$0.0}  & 94.8{\tiny$\pm$0.0}   
			& 90.3{\tiny$\pm$0.3}  & 95.0{\tiny$\pm$0.1}  & 95.0{\tiny$\pm$0.1}  & 95.0{\tiny$\pm$0.1}  
			\\
			
			& DPN$^-$
			& 89.2{\tiny$\pm$0.1} & \textbf{94.5{\tiny$\pm$0.1}}  & \textbf{94.5{\tiny$\pm$0.1}}  & 38.1{\tiny$\pm$0.5}  
			& 92.8{\tiny$\pm$0.1}  & \textbf{96.8{\tiny$\pm$0.1}}  & \textbf{96.8{\tiny$\pm$0.1}}  & 25.4{\tiny$\pm$0.4}  
			& 92.8{\tiny$\pm$0.1}  & \textbf{96.5{\tiny$\pm$0.1}}  & \textbf{96.5{\tiny$\pm$0.1}}  & 31.5{\tiny$\pm$0.4}  
			\\ \hline
		\end{tabular}
	}
	\resizebox{14.0cm}{!}{%
		\begin{tabular}{ll|cccc|cccc|cccc}
			\hline
			& \multicolumn{1}{c|}{OOD} 
			& \multicolumn{4}{c}{CIFAR-10 \cite{db_cifar} }
			& \multicolumn{4}{c}{CIFAR-100  \cite{db_cifar} }
			& \multicolumn{4}{c}{Textures  \cite{db_textures}}  \\
			
			& \multicolumn{1}{c|}{} 
			& Max.P & MI & $\alpha_0$ & D.Ent
			& Max.P & MI & $\alpha_0$ & D.Ent
			& Max.P & MI & $\alpha_0$ & D.Ent\\
			\cline{2-14}
			
			& Baseline
			& 76.9{\tiny$\pm$0.2} & -  & -  & -
			& 73.6{\tiny$\pm$0.2} & - & - & -
			& 70.9{\tiny$\pm$0.2} & -  & - & -
			\\
			
			\multirow{6}{*}{\rotatebox{90}{TIM} } 
			& MCDP
			& 77.4{\tiny$\pm$0.1} & 77.5{\tiny$\pm$0.2} &- &- 
			& 74.0{\tiny$\pm$0.2} & 73.6{\tiny$\pm$0.2}  &- &-
			& 70.3{\tiny$\pm$0.2} & 63.6{\tiny$\pm$0.2}  &- &-
			\\
			
			& DE 
			& 76.9{\tiny$\pm$NA} & 77.7{\tiny$\pm$NA} &- &- 
			& 73.7{\tiny$\pm$NA} & 75.3{\tiny$\pm$NA} &- &-
			& 71.1{\tiny$\pm$NA} & 76.2{\tiny$\pm$NA} &- &-
			\\
			
			
			& OE 
			& 91.3{\tiny$\pm$0.4} & - & - & - 
			& 89.5{\tiny$\pm$0.5} &- &-  &-
			& 95.8{\tiny$\pm$0.3} & - & - & -
			\\
			
			& DPN$_{rev}$ 
			& 85.4{\tiny$\pm$0.7}  & 82.8{\tiny$\pm$1.4}  & 81.9{\tiny$\pm$1.6}  & 85.6{\tiny$\pm$0.9}
			& 84.2{\tiny$\pm$0.8}  & 82.5{\tiny$\pm$1.4}  & 81.7{\tiny$\pm$1.6}  & 85.0{\tiny$\pm$0.9}
			& 90.9{\tiny$\pm$0.3}  & 91.2{\tiny$\pm$0.6}  & 90.6{\tiny$\pm$0.6}  & 92.6{\tiny$\pm$0.3}
			\\
			
			& DPN$^+$ 
			& 99.2{\tiny$\pm$0.0} & 99.7{\tiny$\pm$0.0}  & 99.7{\tiny$\pm$0.0}  & 99.6{\tiny$\pm$0.0}  
			& 98.8{\tiny$\pm$0.0}  & 99.5{\tiny$\pm$0.0}  & 99.5{\tiny$\pm$0.0}  & 99.4{\tiny$\pm$0.0}
			& 96.5{\tiny$\pm$0.1}  & 98.4{\tiny$\pm$0.0}  & 98.4{\tiny$\pm$0.0}  & 98.2{\tiny$\pm$0.0}
			\\
			
			& DPN$^-$ 
			& 99.7{\tiny$\pm$0.0} & \textbf{99.9{\tiny$\pm$0.0}}  & \textbf{99.9{\tiny$\pm$0.0}}  & 3.5{\tiny$\pm$0.1}  
			& 98.7{\tiny$\pm$0.1}  & \textbf{99.6{\tiny$\pm$0.0}}  & \textbf{99.6{\tiny$\pm$0.0}}  & 7.5{\tiny$\pm$0.2}
			& 95.8{\tiny$\pm$0.1}  & \textbf{98.7{\tiny$\pm$0.1}}  & \textbf{98.7{\tiny$\pm$0.1}}  & 19.3{\tiny$\pm$0.4}
			\\ \hline
		\end{tabular}
	}
\end{table}

Tables \ref{table:ood} shows the performance of C10, C100 and TIM classifiers for \textit{OOD detection} task. 
We observe that our DPN$^-$ models consistently outperform the other models using mutual information (MI) measure.
Our  DPN$^-$ models produce sharp multi-modal Dirichlet distributions for OOD examples, leading to higher MI scores compared to the in-domain examples.
In contrast, DPN$^-$ models produce sharp Dirichlet distributions for both in-domain confident predictions and OOD examples. 
Hence, we cannot use D.Ent to distinguish them.
However, in Table \ref{table:gauss_div}, we show that we can combine D.Ent with a total uncertainty measure to distinguish the in-domain and OOD examples.

\begin{table}[h]
	\centering
\caption{ AUROC scores for misclassified image detection.
	Refer to Table \ref{table:misclassification} (Appendix) for additional results by using AUPR scores along with in-domain classification accuracy.
}
\label{table:miss_result}	
	\resizebox{14cm}{!}
	{%
		\begin{tabular}{l|cccc||cccc||cccc}
			\hline
			& \multicolumn{4}{c}{C10} 
			& \multicolumn{4}{c}{C100} 
			& \multicolumn{4}{c}{TIM} \\
			\multicolumn{1}{c|}{} 
			& Max.P & MI & $\alpha_0$ & D.Ent 
			& Max.P & MI & $\alpha_0$ & D.Ent 
			& Max.P & MI & $\alpha_0$ & D.Ent  \\ \cline{2-13}
			Baseline 
			& 93.3{\tiny$\pm$0.1}  &-  &-  &- 
			& 86.8{\tiny$\pm$0.1}  & - & - & - 
			& 86.7{\tiny$\pm$0.0}  &-  &-  &- 
			\\
			
			MCDP 
			& \textbf{93.6{\tiny$\pm$0.2}}  & 93.2{\tiny$\pm$0.1}  &- &-
			& \textbf{87.2{\tiny$\pm$0.0}}  & 83.3{\tiny$\pm$0.3} &- & - 
			& 86.6{\tiny$\pm$0.1}  & 83.3{\tiny$\pm$0.3} &- &- 
			\\
			
			DE 
			& 93.5{\tiny$\pm$NA} & 92.7{\tiny$\pm$NA} &- &-  
			& 87.0{\tiny $\pm$NA} & 83.4{\tiny $\pm$NA} &- &- 
			& \textbf{86.8{\tiny$\pm$NA}} & 83.3{\tiny$\pm$NA} &- &- 
			\\
			

			OE
			& 92.0{\tiny$\pm$0.0} &-  &-  &- 
			& 86.9{\tiny$\pm$0.0} & - & - & -
			& 85.9{\tiny$\pm$0.2} & -  & - & - 
			\\
			
			DPN$_{rev}$ 
			& 89.6{\tiny$\pm$0.1}  & 88.7{\tiny$\pm$0.2}  & 88.7{\tiny$\pm$0.2}  & 89.0{\tiny$\pm$0.2} 
			& 79.3{\tiny$\pm$0.1}  & 73.5{\tiny$\pm$0.1}  & 73.1{\tiny$\pm$0.1}  & 75.7{\tiny$\pm$0.1}
			& 81.9{\tiny$\pm$0.3}  & 72.2{\tiny$\pm$0.7}  & 70.2{\tiny$\pm$0.9}  & 78.3{\tiny$\pm$0.3}  
			\\
			
			DPN$^+$ 
			& 92.2{\tiny$\pm$0.3}  & 90.3{\tiny$\pm$0.1}  & 90.3{\tiny$\pm$0.1}  & 90.5{\tiny$\pm$0.2}
			& 86.5{\tiny$\pm$0.1}  & 81.2{\tiny$\pm$0.0}  & 81.3{\tiny$\pm$0.0}  & 81.9{\tiny$\pm$0.1}
			& 85.7{\tiny$\pm$0.2} & 78.3{\tiny$\pm$0.4}  & 78.7{\tiny$\pm$0.5}  & 79.7{\tiny$\pm$0.2}
			\\
			
			DPN$^-$  
			& 92.6{\tiny$\pm$0.1}  & 89.9{\tiny$\pm$0.0}  & 89.9{\tiny$\pm$0.0}  & 66.2{\tiny$\pm$0.7}
			& 86.4{\tiny$\pm$0.1}  & 82.3{\tiny$\pm$0.0}  & 82.3{\tiny$\pm$0.0}  & 81.7{\tiny$\pm$0.1}
			& 85.4{\tiny$\pm$0.1}  & 79.1{\tiny$\pm$0.5}  & 79.4{\tiny$\pm$0.4}  & 79.9{\tiny$\pm$0.2}
			\\
			\hline
		\end{tabular}
	}
\end{table}

Table \ref{table:miss_result} presents the results for in-domain \textit{misclassification detection}. 
We can see that our proposed DPN$^-$ models achieve comparable performance as the existing methods.
It is interesting to note that, all the DPN models produce comparable AUROC scores using the distributional uncertainty measures as the total uncertainty measure.
This supports our assertion that \textit{in the presence of data uncertainty, DPN models tend to produce flatter and diverse Dirichlet distributions with smaller precisions for misclassified examples, compared to the confident predictions.}
Hence, we should aim to produce sharp multi-modal Dirichlet distributions for OOD examples to keep them distinguishable from the in-domain examples.
In Appendix \ref{sec:rmsc_app}, we also demonstrate that our DPN$^-$ models  improve calibration performance for classification \cite{calib_icml_2017,oe_iclr_2019}.


AUROC (and also AUPR) scores, in Table \ref{table:ood}, only provide a relative measure of separation, while providing no information about how different these uncertainty values are for in-domain and OOD examples \cite{roc_psychology_2014}.
An OOD detection model should aim to maximize the margin of uncertainty values produced for the OOD examples from the in-domain examples to separate them efficiently.
We can measure the ``gap'' of uncertainty values for in-domain and OOD examples by measuring the divergence of their distributions for uncertainty values produced by the models.

\begin{wrapfigure}{r}{7.5cm}
	\centering
	\vspace{-1.0em}
	\caption{ Illustrating the distribution of uncertainty values for DPN$^-$ and other DPN models.
	We normalize the scores for better visualization.}	
	\begin{subfigure}[t]{0.49\linewidth}
		\captionsetup{justification=centering}
		\caption*{\small (a) DPN$^-$}		
		\includegraphics[width=1\linewidth, height=90pt]{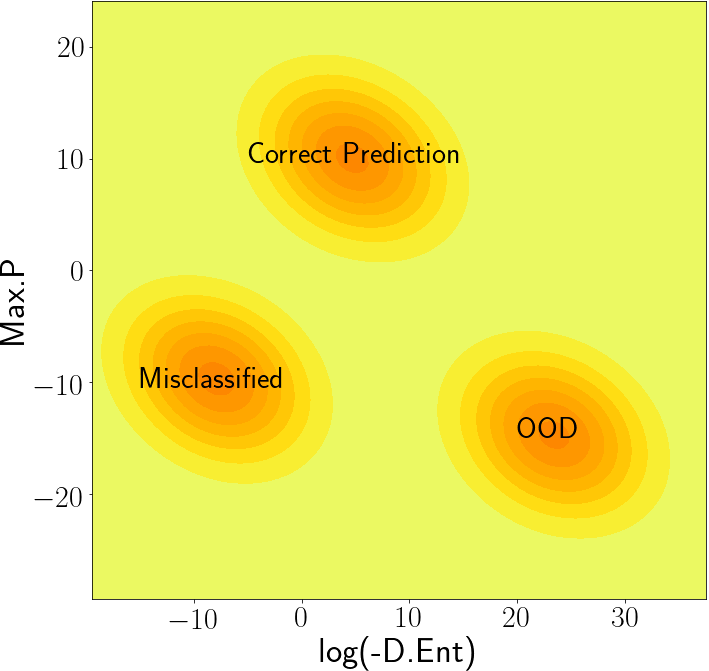}
	\end{subfigure}
	\begin{subfigure}[t]{0.49\linewidth}
		\captionsetup{justification=centering}
		\caption*{\small (b) Other DPN models}
		\includegraphics[width=1\linewidth, height=90pt]{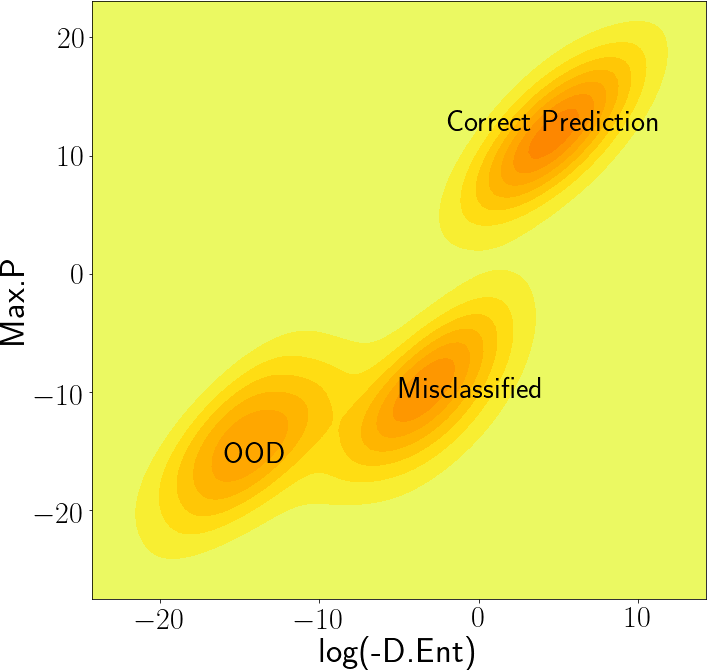}		
	\end{subfigure}
	\vspace{-0.5em}
	\label{fig:uncertainty_dist}
\end{wrapfigure}

DPN$^-$ models produce sharp Dirichlet distributions for both OOD examples and in-domain confident predictions, while flat Dirichlet distributions for misclassified examples.
Hence, we can combine D.Ent along with a total uncertainty measure, Max.P, to robustly distinguish them.
Note that D.Ent ranges from $(-\infty, \infty)$ and often produce large negative values. 
Hence, we use $\log$(-D.Ent).
We separately consider Max.P and $\log$(-D.Ent) scores for the in-domain correctly classified, misclassified, and OOD examples.
We fit three different bivariate Gaussian distributions on these values.
We then compute the KL-divergence of the distributions of uncertainty values for in-domain confidently predicted and misclassified examples to the OOD examples to measure their separability.
Figure \ref{fig:uncertainty_dist} illustrates the desired behavior of our DPN$^-$, compared to the other DPN models.

\begin{table}[h]
	\vspace*{-0.5em}
\caption{ KL-divergence scores from the distribution of uncertainty values of missclassified and correctly predicted examples to the OOD examples.
	See Table \ref{table:kl_div_app} (Appendix) for additional results.
}
\label{table:gauss_div}	
	\centering
	\resizebox{14cm}{!}{%
		\begin{tabular}{l|cc|cc||cc|cc||cc|cc}
			\hline
			& \multicolumn{4}{c}{C10} 
			& \multicolumn{4}{c}{C100} 
			& \multicolumn{4}{c}{TIM} \\
			\multicolumn{1}{c|}{OOD} & 
			\multicolumn{2}{c}{Tiny \cite{db_tiny} } & \multicolumn{2}{c||}{LSUN \cite{db_lsun} } & 
			\multicolumn{2}{c}{Tiny \cite{db_tiny} } & \multicolumn{2}{c||}{LSUN \cite{db_lsun} } & 
			\multicolumn{2}{c}{CIFAR-10 \cite{db_cifar}} & \multicolumn{2}{c}{CIFAR-100  \cite{db_cifar} } 
			\\ \cline{2-13}
			\multicolumn{1}{c|}{} & 
			Miss & Correct & Miss & Correct 
			& Miss & Correct & Miss & Correct 
			& Miss & Correct & Miss & Correct \\
			DPN$_{rev}$ 
			& 1.4{\tiny$\pm$0.2}  & 12.6{\tiny$\pm$0.8} 
			& 2.1{\tiny$\pm$0.3}  & 13.9{\tiny$\pm$1.0} 
			& 1.8{\tiny$\pm$0.0}  & 6.6{\tiny$\pm$0.1} 
			& 2.6{\tiny$\pm$0.0}  & 8.6{\tiny$\pm$0.1} 
			& 9.0{\tiny$\pm$0.6}  & 9.2{\tiny$\pm$1.0} 
			& 2.3{\tiny$\pm$0.4}  & 3.2{\tiny$\pm$0.5}   
			\\
			
			DPN$^+$ 
			& 1.5{\tiny$\pm$0.2}  & 12.1{\tiny$\pm$1.9} 
			& 2.0{\tiny$\pm$0.2}  & 12.7{\tiny$\pm$2.0} 
			& 1.8{\tiny$\pm$0.0}  & 7.2{\tiny$\pm$0.2} 
			& 2.6{\tiny$\pm$0.0}  & 9.1{\tiny$\pm$0.2} 
			& 21.1{\tiny$\pm$3.4}  & 27.1{\tiny$\pm$3.9} 
			& 17.7{\tiny$\pm$3.0}  & 23.5{\tiny$\pm$3.5}   
			\\
			
			DPN$^-$
			& \textbf{2.1{\tiny$\pm$0.1}}  & \textbf{20.7{\tiny$\pm$1.2}} 
			& \textbf{2.4{\tiny$\pm$0.1}}  & \textbf{22.5{\tiny$\pm$1.3}} 
			& \textbf{44.5{\tiny$\pm$2.8}}  & \textbf{244.2{\tiny$\pm$25.2}}
			& \textbf{49.9{\tiny$\pm$3.5}}  & \textbf{272.0{\tiny$\pm$29.3}} 
			& \textbf{729.8{\tiny$\pm$12.4}}  & \textbf{1360.4{\tiny$\pm$94.5}} 
			& \textbf{664.9{\tiny$\pm$11.6}}  & \textbf{1241.4{\tiny$\pm$79.8}}   
			\\
			\hline
		\end{tabular}
	}
\end{table}
%
%

In Table \ref{table:gauss_div}, the significantly higher KL-divergence for our DPN$^-$ models indicate that by combining Max.P with D.Ent measure, we can easily distinguish the OOD examples from the in-domain examples.
Further, as we consider classification tasks with a larger number of classes, our DPN$^-$ produces more number of fractional concentration parameters for each class for the OOD examples. 
It further increases the $\log$(-D.Ent) values, leading to maximizing the ``gaps" between OOD examples from both in-domain confident predictions as well as misclassified examples.
\section{Conclusion}

The existing formulation for DPN models often lead to indistinguishable representations between in-domain examples with high data uncertainty among multiple classes and OOD examples.
In this work, we have proposed a novel loss function for DPN models that maximizes the representation gap between in-domain and OOD examples.
Experiments on benchmark datasets demonstrate that our proposed approach effectively distinguishes the distributional uncertainty from other uncertainty types and outperforms the existing OOD detection models.

\section{Broader Impact}
Despite the impeccable success of deep neural network (DNN)-based models in various real-world applications, they often produce incorrect predictions without proving any warning for the users.
It raises the question of how much can we trust these models and whether it is safe to use them for sensitive real-world applications such as medical diagnosis, self-driving cars or to make financial decisions.

In this paper, we aim to robustly identify the source of uncertainty in the prediction of a DNN based model for classification tasks.
Identifying the source of uncertainty in the prediction would allow manual intervention in an informed way and make an AI system more reliable for real-world applications.
In particular, we address a shortcoming of the existing techniques and propose a novel solution to improve the detection of anomalous out-of-distribution examples for a classification model.


\section*{Acknowledgement}
This research is supported by the National Research Foundation, Singapore under its
AI Singapore Programme (AISG Award No: AISG-GC-2019-001). 
Any opinions, findings and conclusions or recommendations expressed in this material are those of the authors and do not reflect the views of National Research Foundation, Singapore

\bibliography{uncertainty}
\bibliographystyle{unsrt}

\newpage
\appendix

\begin{center}
{\LARGE{Appendix}}
\end{center}

\normalsize
\textbf{\textsf{Organization:}}
We organize the appendix as follows:

\begin{enumerate}

\item We present a set of ablation studies for our models in Section \ref{sec:ablation}.

\item Section \ref{sec:toy_app} provides the implementation details for our experiments on synthetic datasets. 

Section \ref{sec:benchmark_app} provides the experimental setup, implementation details of our models, and competitive models along with the description of the OOD test datasets for our experiments on the benchmark image classification datasets.

Section \ref{sec:rmsc_app} presents a comparative study for confidence calibration performance of different models.

\item The expressions for differential entropy, mutual information of a Dirichlet distribution, and the KL Divergence between two Gaussian distributions are provided in Section \ref{sec:derivations}.

\item The extended results (mean $\pm$ standard deviation of 3 models) for the benchmark image classification datasets are provided in Section \ref{sec:extended} from Table \ref{table:misclassification} to Table \ref{table:extended_result_tim}.
\end{enumerate}

\section{Ablation Studies}
\label{sec:ablation}
\subsection{Different choices for $\lambda_{in}$ and $\lambda_{out}$}
\label{sec:lambda_0}
Choosing both $\lambda_{in}$ and $\lambda_{out}$ to $0$ lead Eqn \ref{eq:l_total_reg} to the same loss function as non-Bayesian outlier exposure (OE) framework \cite{oe_iclr_2019}, while loosing control over the  precision of the output Dirichlet distributions.
In contrast, setting either $\lambda_{in}$ or $\lambda_{out}$ to $0$ loses control over the precision for in-domain or OOD examples respectively.
We train two additional DPN models, denoted as DPN$_{\{0, -0.5\}}$ and DPN$_{\{0.5,0\}}$ for C100 classification tasks to investigate the choices of these hyper-parameters.
DPN$_{\{0, -0.5\}}$ is trained using  $\lambda_{in} = 0$ and $\lambda_{out} = -0.5$.
DPN$_{\{0.5,0\}}$ is trained using $\lambda_{in} = 0.5$ and $\lambda_{out} = 0$.
In Table \ref{table:ablation_lambda}, we present their comparative performance with DPN$^-$.

\begin{table}[h]
\centering
\caption{AUROC scores for OOD image detection of our DPN$^-$ models using different values of $\lambda_{in}$ and $\lambda_{out}$ for C100 classification task.
	We report (mean $\pm$ standard deviation) values of three runs.}
\vspace{0.5em}
\label{table:ablation_lambda}
\resizebox{14.0cm}{!}{%
\begin{tabular}{l|cccc|cccc|cccc}
\hline
OOD 
& \multicolumn{4}{c}{Tiny}
& \multicolumn{4}{c}{STL-10}
& \multicolumn{4}{c}{LSUN}  \\
\multicolumn{1}{c|}{} 
& Max.P & MI & $\alpha_0$ & D.Ent 
& Max.P & MI & $\alpha_0$ & D.Ent 
& Max.P & MI & $\alpha_0$ & D.Ent 
\\ \hline
& \multicolumn{4}{c|}{} & \multicolumn{4}{c|}{} & \multicolumn{4}{c}{}
\\			
DPN$_{\{0, -0.5\}}$ 
& 84.6{\tiny$\pm$0.0}  & 91.1{\tiny$\pm$0.1}  & 91.6{\tiny$\pm$0.2}  & 21.7{\tiny$\pm$0.7} 
& 90.2{\tiny$\pm$0.1}  & 95.0{\tiny$\pm$0.0}  & 95.7{\tiny$\pm$0.1}  & 12.1{\tiny$\pm$0.3}
& 88.5{\tiny$\pm$0.2}  & 93.8{\tiny$\pm$0.1}  & 94.2{\tiny$\pm$0.1}  & 17.5{\tiny$\pm$0.3}  

\\
			
DPN$_{\{0.5, 0\}}$ 
& 89.2{\tiny$\pm$0.4}  & 93.6{\tiny$\pm$0.1}  & 93.6{\tiny$\pm$0.1}  & 93.6{\tiny$\pm$0.1}  
& 92.1{\tiny$\pm$0.5}  & 96.2{\tiny$\pm$0.3}  & 96.2{\tiny$\pm$0.3}  & 96.0{\tiny$\pm$0.2}
& 92.0{\tiny$\pm$0.3}  & 95.9{\tiny$\pm$0.1}  & 95.9{\tiny$\pm$0.1}  & 95.9{\tiny$\pm$0.1}

\\
			
DPN$^-$ 
& 89.2{\tiny$\pm$0.1} & \textbf{94.5{\tiny$\pm$0.1}}  & \textbf{94.5{\tiny$\pm$0.1}}  & 38.1{\tiny$\pm$0.5}  
& 92.8{\tiny$\pm$0.1}  & \textbf{96.8{\tiny$\pm$0.1}}  & \textbf{96.8{\tiny$\pm$0.1}}  & 25.4{\tiny$\pm$0.4}  
& 92.8{\tiny$\pm$0.1}  & \textbf{96.5{\tiny$\pm$0.1}}  & \textbf{96.5{\tiny$\pm$0.1}}  & 31.5{\tiny$\pm$0.4}  

\\
\hline
\end{tabular}
}
\end{table}
\textbf{Analyzing DPN$_{\{0, -0.5\}}$: } 
The choice of only $\lambda_{in} = 0$ in Eqn. \ref{eq:l_total_reg} does not enforce the DPN to produce larger concentration parameters for the in-domain examples. 
It only learns to produce fractional (i.e <1) concentration parameters for OOD examples, leading to produce sharper multi-modal Dirichlet distributions for OOD examples.

However, now the network can produce fractional (i.e <1) concentration parameters even for in-domain examples as well.
This leads to inappropriately interpolate the concentration parameters in the boundary of in-domain and OOD regions.
As a result, it leads to degrading the OOD detection performance.
We can see that, similar to DPN$^-$ models, DPN$_{\{0, -0.5\}}$ models also produce lower AUROC scores for D.Ent.
This indicates that the choice of $\lambda_{in} = 0, \lambda_{out} < 0$ leads to produce sharp multi-modal Dirichlet distributions for OOD examples.
However, their overall OOD detection performance degrade compare to DPN$^-$ models.

\textbf{Analyzing DPN$_{\{0.5, 0\}}$: }
On the other hand, DPN$_{\{0.5,0\}}$ demonstrates similar property as DPN$^+$.
In this case, the network produces flatter Dirichlet distributions for OOD examples compare to the in-domain examples.
As we can see that DPN$_{\{0.5,0\}}$ produces high AUROC scores for D.Ent measure.
However, as before, it does not address the issue of efficiently maximizing the \textit{`representational gap'} between in-domain and OOD examples.
We can see in Table \ref{table:ablation_lambda}, DPN$_{\{0.5,0\}}$ cannot exceed the OOD detection performance of DPN$^-$ models, similar to the DPN$^+$ models.
\subsection{A different choice of $ \beta_{out}$ for RKL loss}
\label{sec:rkl_fraction}
In section \ref{sec:proposed}, we explain that choosing fractional values for target concentration parameters, $\bm \beta_{out}$ for RKL loss \cite{dpn2_nips_2019} does not guarantee to produce fractional concentration parameters for OOD examples (see Eq. \ref{eq:rkl_ood}).
Here, we investigate this by choosing the target concentration parameters to $0.1$ for all classes for OOD training examples.
For in-domain training examples, we set the target concentration parameters as $100$ for the correct class and $1$ for the incorrect classes.
We denote it as DPN$^{0.1}_{rev}$. 

In Table \ref{table:ablation_rkl}, we compare their OOD detection performance with the standard DPN$_{rev}$ models where the target concentration parameters for OOD examples are set to $1$ for all classes.
We observe that the performance of DPN$^{0.1}_{rev}$ models produce lower AUROC scores for D.Ent measures, while their overall performance degrade compare to the standard DPN$_{rev}$ models.
This is because DPN$^{0.1}_{rev}$ models often produce both greater than and less than $1$ values of concentration parameters of OOD examples, 
that leads to uni-modal Dirichlet distributions, instead of a multi-modal Dirichlet.
This representation is often similar to the in-domain examples.
Hence, it becomes even more difficult to distinguish the in-domain and OOD examples for DPN$^{0.1}_{rev}$ models, which lead to degrade their overall performance.

\begin{table}[htbp]
\centering
\caption{AUROC scores OOD image detection results for DPN models using RKL loss function \cite{dpn2_nips_2019} with different choices of hyper-parameters for C100 classification task.
We report (mean $\pm$ standard deviation) values of three runs.}
\vspace{0.5em}
\label{table:ablation_rkl}
\resizebox{14.0cm}{!}{%
\begin{tabular}{l|cccc|cccc|cccc}
\hline
OOD 
& \multicolumn{4}{c}{Tiny}
& \multicolumn{4}{c}{STL-10}
& \multicolumn{4}{c}{LSUN}  \\

\multicolumn{1}{c|}{} 
& Max.P & MI & $\alpha_0$ & D.Ent 
& Max.P & MI & $\alpha_0$ & D.Ent 
& Max.P & MI & $\alpha_0$ & D.Ent 
\\ \hline
& \multicolumn{4}{c|}{} & \multicolumn{4}{c|}{} & \multicolumn{4}{c}{}
\\

DPN$_{rev}^{0.1}$ 
& 74.9{\tiny$\pm$0.2}  & 80.4{\tiny$\pm$0.2}  & 80.7{\tiny$\pm$0.2}  & 48.1{\tiny$\pm$0.1} 
& 78.1{\tiny$\pm$0.1}  & 84.3{\tiny$\pm$0.1}  & 84.7{\tiny$\pm$0.1}  & 32.8{\tiny$\pm$0.2}
& 76.7{\tiny$\pm$0.1}  & 82.2{\tiny$\pm$0.0}  & 82.5{\tiny$\pm$0.1}  & 49.3{\tiny$\pm$0.2}  

\\

DPN$_{rev}$ 
& 81.2{\tiny$\pm$0.2} & \textbf{83.8{\tiny$\pm$0.1}} & \textbf{83.8{\tiny$\pm$0.1}} & 83.5{\tiny$\pm$0.1}  
& 87.2{\tiny$\pm$0.1} & \textbf{89.3{\tiny$\pm$0.1}} & \textbf{89.3{\tiny$\pm$0.1}} & 89.0{\tiny$\pm$0.1}   
& 86.7{\tiny$\pm$0.0} & \textbf{89.3{\tiny$\pm$0.1}} & \textbf{89.3{\tiny$\pm$0.1}} & 88.9{\tiny$\pm$0.1}  

\\
\hline
\end{tabular}
}
\end{table}
\subsection{A Binary Classifier for OOD Detection}
In this work, we show that in the presence of high data uncertainty, the existing OOD detectors often lead to the same representation for in-domain examples as the OOD examples.
Hence, one can simply think of training a binary classifier using in-domain and OOD training examples as two different classes to distinguish between in-domain examples and OOD examples.
Since it does not need to classify the in-domain examples among multiple classes, it would not suffer from data uncertainty and should automatically solve the problem.
However, note that such a binary classifier only learns to produce sharp categorical distributions for these training examples.
Hence, given an unknown OOD test example, it does not necessarily produce sharp categorical distribution for the OOD class.

For example, for our experiment on C10 classification task, we use CIFAR-10 training images as the in-domain training set and CIFAR-100 training examples the OOD training set.
In contrast, for C100 classification task, we use CIFAR-100 training images as the in-domain training set and CIFAR-10 training examples as the OOD training set.
Hence, given an OOD test example from TIM dataset, if the binary classifier for C10 produces a higher probability score for the OOD class, it is expected to produce a lower probability score for the OOD class for C100 classification task.
In contrast, our DPN$^-$ models are explicitly trained to produce multi-modal Dirichlet distributions for an unknown test example, whenever it does not `fit' into the in-domain class-labels.

\begin{table}[h]
\caption{OOD image detection results of the binary classifiers compare to our DPN$^-$ models for C10 and C100 classification task.
	We report (mean $\pm$ standard deviation) values of three runs.}
\vspace{0.5em}
\label{table:ablation_1}
\centering
	\resizebox{11.0cm}{!}{%
		\begin{tabular}{l|cc|cc|cc|cc}
		\hline
			\multirow{3}{*}{OOD} 
			& \multicolumn{4}{c|}{C-10 Classification}
			& \multicolumn{4}{c}{C-100 classification}  \\
			& \multicolumn{2}{c}{AUROC}
			& \multicolumn{2}{c|}{AUPR}
			& \multicolumn{2}{c}{AUROC}
			& \multicolumn{2}{c}{AUPR}  			
			\\
			\multicolumn{1}{c|}{} 
			& \multicolumn{1}{c}{Binary} & \multicolumn{1}{c|}{DPN$^-$}
			& \multicolumn{1}{c}{Binary} & \multicolumn{1}{c|}{DPN$^-$}
			& \multicolumn{1}{c}{Binary} & \multicolumn{1}{c|}{DPN$^-$}
			& \multicolumn{1}{c}{Binary} & \multicolumn{1}{c}{DPN$^-$}
			\\ \hline
			
			Tiny 
			& \textbf{99.0{\tiny$\pm$0.2}} & \textbf{99.0{\tiny$\pm$0.1}}	
			& \textbf{99.1{\tiny$\pm$0.1}} & \textbf{99.1{\tiny$\pm$0.1}}
			& 84.8{\tiny$\pm$1.2} 		& \textbf{94.5{\tiny$\pm$0.1}}
			& 87.9{\tiny$\pm$0.7}		& \textbf{95.1{\tiny$\pm$0.1}}	
			\\
			
			STL-10 
			& \textbf{93.7{\tiny$\pm$0.9}} & 85.3{\tiny$\pm$0.5}
			& \textbf{94.0{\tiny$\pm$0.7}} & 85.2{\tiny$\pm$0.6}
			& 90.7{\tiny$\pm$0.9} & \textbf{96.8{\tiny$\pm$0.1}}
			& 91.0{\tiny$\pm$0.6} & \textbf{96.7{\tiny$\pm$0.1}}		
			\\			
			
			LSUN 
			& 98.9{\tiny$\pm$0.2} & \textbf{99.3{\tiny$\pm$0.0}}
			& 98.9{\tiny$\pm$0.2} & \textbf{99.3{\tiny$\pm$0.1}}
			& 91.2{\tiny$\pm$1.2} & \textbf{96.5{\tiny$\pm$0.1}}
			& 93.0{\tiny$\pm$0.8} & \textbf{97.0{\tiny$\pm$0.1}}
			\\
			
			Places365
			& 98.8{\tiny$\pm$0.2} & \textbf{98.9{\tiny$\pm$0.1}}
			& \textbf{99.7{\tiny$\pm$0.0}} & \textbf{99.7{\tiny$\pm$0.0}}
			& 88.2{\tiny$\pm$1.3} & \textbf{94.5{\tiny$\pm$0.1}}
			& 96.7{\tiny$\pm$0.3} & \textbf{98.4{\tiny$\pm$0.0}}
			\\
			
			Textures
			& \textbf{99.7{\tiny$\pm$0.1}} & \textbf{99.7{\tiny$\pm$0.0}}
			& \textbf{99.6{\tiny$\pm$0.1}} & 99.4{\tiny$\pm$0.1}
			& 65.4{\tiny$\pm$1.5} & \textbf{85.2{\tiny$\pm$0.1}} 
			& 65.2{\tiny$\pm$0.9} & \textbf{78.9{\tiny$\pm$0.2}}
			\\ 
			\hline
		\end{tabular}
}
\end{table}

In Table \ref{table:ablation_1}, we compare the OOD detection performance of such binary classifiers with our DPN$^-$ models for C10 and C100.
For the binary classifier, we consider the probability score of the ``in-domain" class as their uncertainty metric. 
For our DPN$^-$ models, we select the best AUROC and AUPR values from Table \ref{table:extended_result_cifar10} and Table \ref{table:extended_result_cifar100} for C10 and C100 classification tasks respectively..
We can see that, while the binary classifier often out-performs our DPN$^-$ models, it does not necessarily provide the upper bound for the OOD detection tasks.
In practice, since we do not know the characteristics of the OOD test examples, it may not be suitable to use a binary classifier for OOD detection tasks.

\begin{figure}[ht]
	\vspace{-0.5em}
	\caption{
			Visualization and understanding the desired characteristics of different DPN models.
			We visualize the uncertainty measures for different data-points for DPN$_{rev}$, DPN$^+$ and DPN$^-$.
	}
	\label{fig:toy_experiment_app}
	\centering
	\begin{minipage}{1\textwidth}
		\centering
		\begin{subfigure}{\linewidth}
			\centering 			
			\includegraphics[width=0.3\linewidth,height=120pt]{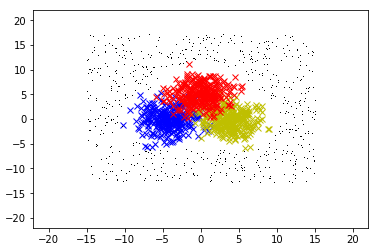}
			\caption{\small In-domain \& OOD training examples}
			\label{fig:sub1}
		\end{subfigure}
	\end{minipage}
	
	\begin{minipage}{\textwidth}
		\begin{subfigure}{\linewidth}
			\centering	
			\includegraphics[width=1\linewidth,height=70pt]{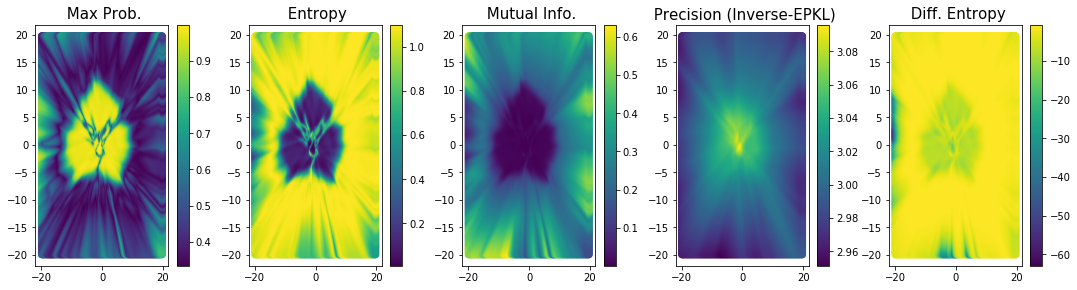}
			\caption{\small DPN$_{rev}$}
			\label{fig:toy_d_rev}
		\end{subfigure}
		\begin{subfigure}{\linewidth}
			\centering
			\includegraphics[width=1\linewidth,height=70pt]{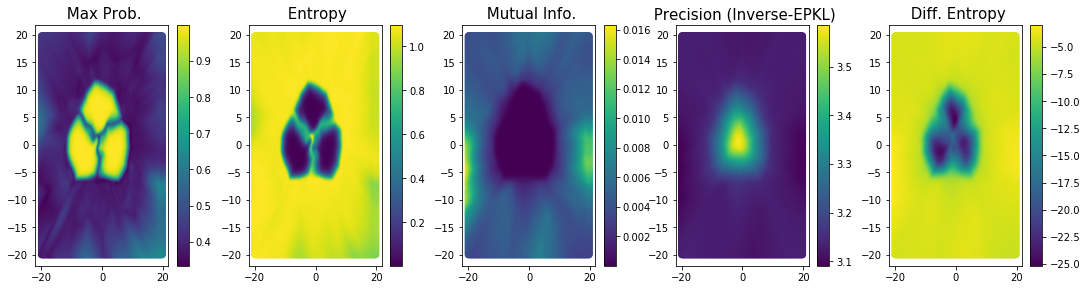}
			\caption{\small DPN$^+$}
			\label{fig:toy_d_+}
		\end{subfigure}\\
		\begin{subfigure}{\linewidth}
			\centering
			\includegraphics[width=1\linewidth,height=70pt]{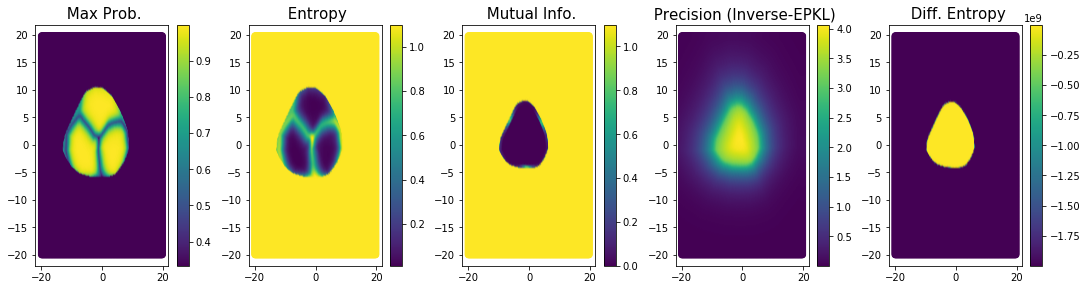}
			\caption{\small DPN$^-$}
			\label{fig:toy_d_-}
		\end{subfigure}
	\end{minipage}%
\end{figure}

\section{Implementation Details and Extended Results}
\label{sec:implementation_appendix}

\subsection{Synthetic Datasets}
\label{sec:toy_app}
The three classes of our synthetic dataset are constructed by sampling from 
three different isotropic Gaussian distributions with means of
$(-4,0)$, $(4,0)$ and $(0,5)$ and isotropic variances of $\sigma=4$.
We sample $200$ training data points from each distribution for each class.
We also draw $600$ samples for OOD training examples from a uniform distribution of $\mathcal{U}([-15, 15], [-13, 17])$, outside the Gaussian distributions.

\paragraph{Hyper-parameters. }
We train a neural network with 2 hidden layers with 125 nodes each and $relu$ activation function.
The training code is provided along with the supplementary materials.

We use a neural network with 2 hidden layers with 50 nodes each and ReLU activation function.
We set $\gamma=1.0$ in the overall loss function.
We have two DPN models. 
Our first DPN model, DPN$^+$, is trained using a positive $\lambda_{out} = \frac{1}{\#class} + 0.5$ and   $\lambda_{in} = 1.5$.
The second DPN model, DPN$^-$, is trained using a negative $\lambda_{out} =\frac{1}{\#class} - 0.5$, and $\lambda_{in} = 0.5$.
Our DPN$^+$ and DPN$^-$ models are trained using the SGD optimizer.

We also train a DPN$_{rev}$ model by using RKL loss \cite{dpn2_nips_2019}. 
We set the concentration hyper-parameters as follows:
for in-domain training examples, we set the concentration parameters to $1e2$ for the correct class and $1$ for the incorrect classes.
For the OOD training examples, we set the concentration parameters as $1$ for all classes. 
We train the DPN$_{rev}$ model using ADAM optimizer \cite{adam_2014}.
Notably, we could not train the DPN$_{rev}$ model using SGD optimization due to their complex RKL loss function.

\paragraph{Additional Results. }
We visualize the uncertainty measures of different data points for DPN$_{rev}$ along with our DPN$^+$ and DPN$^-$ models in Figure \ref{fig:toy_experiment_app}(b), \ref{fig:toy_experiment_app}(c) and \ref{fig:toy_experiment_app}(d) respectively. 
We observe very similar characteristics for DPN$_{rev}$ and our DPN$^+$.
In contrast, our DPN$^-$ produces much sharper boundaries to distinguish the in-domain and OOD examples using distributional uncertainty measure i.e mutual information and precision (or inverse-EPKL) to demonstrate its superiority.

We also present the results for $entropy$ of categorical posterior distributions, $\mathcal{H}[ p(\omega_c| \bm{x}^*, D)]$, for different data points.
This is a total uncertainty measure as it is derived from the expected predictive categorical distribution, $p(\omega_c| \bm{x}^*, D)$ i.e by marginalizing $\bm \mu$ and $\bm \theta$ in Eq. \ref{eq:dpn}.

\subsection{Benchmark Image Classification Datasets}
\label{sec:benchmark_app}
We use the VGG-16 network for C10, C100, and TIM classification tasks.
For C10, we use CIFAR-10 training images ($50,000$ images) as our in-domain training data 
and CIFAR-100 training images ($50,000$ images) as our OOD training data.

For C-100, we use CIFAR-100 training images ($50,000$ images) as our in-domain training data 
and CIFAR-10 training images ($50,000$ images) as our OOD training data.

For TIM, we use TIM training images (100,000 images) as our in-domain training data and ImageNet-25K images ($25,000$ images) as our OOD training data.
ImageNet-25K is obtained by randomly selecting $25,000$ images from the ImageNet dataset \cite{db_imagenet}.

\paragraph{Hyper-parameters for our DPN$^+$ \& DPN$^-$ models. }
Similar to \cite{dpn_nips_2018,dpn2_nips_2019,oe_iclr_2019}, we do not need to tune any hyper-parameters during testing.
In other words, the OOD test examples remain unknown to our DPN classifiers, as in a real-world scenario.
We set $\gamma = 0.5$ for our loss function in Eqn. \ref{eq:l_total_reg}, as applied in \cite{oe_iclr_2019}. 

We train two different DPN models for each classification task, using both positive and negative values for $\lambda_{out}$ to analyze the effect of flat Dirichlet distributions and sharp Dirichlet distributions across the edges of the simplex respectively for the OOD examples.
Note that we cannot choose arbitrarily large values for $\lambda_{in}$ and $\lambda_{out}$ as it would degrade the in-domain classification accuracy of the models.
For our experiments in the main paper, we select the hyper-parameters as follows:
Our first model, DPN$^+$ is trained with positive 
$\lambda_{out} = \frac{1}{\#class} +0.5$ and   $\lambda_{in} = 1.5$.
The second model, DPN$^-$ is trained with negative 
$\lambda_{out} =\frac{1}{\#class} - 0.5$, and $\lambda_{in} = 0.5$.

\begin{table}[h]
\centering
\caption{Details of Train and Test Datasets used for the different classifiers.}
\vspace{0.5em}
\label{table:details_1}
\resizebox{14.0cm}{!}{%
\begin{tabular}{c|c|c|cc|cc}
\hline
\multirow{2}{*}{Classifier} & \multirow{2}{*}{Input} & \multirow{2}{*}{\#Classes} & \multicolumn{2}{c}{Training Datasets} & \multicolumn{2}{c}{ Test Datasets} \\
 & & & In-Domain & \multicolumn{1}{c}{OOD} &  In-Domain & \multicolumn{1}{c}{OOD}  \\ \hline
& & & & & & \\ 
C10 & $32 \times 32$ & 10 
& \begin{tabular}[c]{@{}c@{}}CIFAR-10 Training Set\\ (50,000 images)\end{tabular} 
& \begin{tabular}[c]{@{}c@{}}CIFAR-100 Training Set\\ (50,000 images)\end{tabular} 
& \begin{tabular}[c]{@{}c@{}}CIFAR-10 Test Images  \\(10,000 Images) \end{tabular} & \begin{tabular}[c]{@{}c@{}} Tiny, STL-10,\\ LSUN etc. \end{tabular} 
\\ \hline
& & & & & & \\ 
C100 & $32 \times 32$ & 100 
& \begin{tabular}[c]{@{}c@{}}CIFAR-100 Training Set\\ (50,000 images)\end{tabular} 
& \begin{tabular}[c]{@{}c@{}}CIFAR-10 Training Set\\ (50,000 images)\end{tabular} 
& \begin{tabular}[c]{@{}c@{}}CIFAR-100 Test Set \\ (10,000 Images) \end{tabular} 
& \begin{tabular}[c]{@{}c@{}} Tiny, STL-10,\\ LSUN etc.  \end{tabular} \\ \hline
& & & & & & \\ 
TIM & $64 \times 64$ & 200 
& \begin{tabular}[c]{@{}c@{}}TinyImageNet Training Set\\ (100,000 images)\end{tabular} 
& \begin{tabular}[c]{@{}c@{}}ImageNet-25K\\ (25,000 randomly sampled \\ images from ImageNet)\end{tabular} 
& \begin{tabular}[c]{@{}c@{}} TIM Test Images \\ (10,000 Images) \end{tabular}
& \begin{tabular}[c]{@{}c@{}} CIFAR-10, \\ CIFAR-100, \\ Textures etc.\end{tabular}
\\ \hline
\end{tabular}%
}
\end{table}
\paragraph{Competitive Systems: Implementations and Additional Discussions. }
We compare the performance of our models with standard DNN as baseline model \cite{baseline_iclr_2017}, Bayesian Monte-Carlo dropout (MCDP) \cite{mcDrop_icml_gal16},
deep-ensemble models (DE) \cite{ensemble_nips_2017} and 
evidential deep learning (EDL) \cite{edl_nips_2018}, 
DPN$_{fwd}$ and DPN$_{rev}$ using the forward and reverse KL-divergence loss function proposed in \cite{dpn_nips_2018} and \cite{dpn2_nips_2019} respectively, non-Bayesian frameworks such as outlier exposure (OE) \cite{oe_iclr_2019}.
We use the same architecture as our DPN models for the other competitive models.

\textbf{MCDP models: }
For MCDP models, we use the standard DNN (baseline) model with randomly dropping the nodes during test time.
The predictive categorical distributions are obtained by averaging the outputs for $10$ iterations.
For OE, DPN$_{fwd}$ and DPN$_{rev}$ models, we use the same setup as applied for our DPN models (See Table \ref{table:details_1}).
Table \ref{table:misclassification} presents the classification accuracies achieved by different models for different classification tasks, along with their performance for misclassified example detection.
For DE, we use an ensemble of three baseline DNN models for our experiments.

\textbf{EDL models:}
Similar to DPN models, EDL also produces Dirichlet distribution for the input examples \cite{edl_nips_2018}.
Unlike DPN, EDL applies ReLU activation, instead of the exponential function (Eqn. \ref{eq:softmax_dpn}), to induce non-negative constraints to produce the concentration parameters of their output Dirichlet distributions. 

EDL models are trained using the in-domain examples.
Their network is trained using a loss function that explicitly maximizes the concentration parameter of the correct class while minimizing the overall precision of the Dirichlet for each in-domain training examples.
For C10 classifiers, we train the EDL models from the scratch. 
For C100 and TIM classifiers, we initialize the EDL models using the pre-trained Baseline models to achieve competitive in-domain classification accuracy.
Next, we replace the soft-max using ReLU activation and retrain the models using the proposed loss function for EDL.

Since both DPN and EDL models output Dirichlet distribution for an input example, we can define the same uncertainty measures as the DPN models. 
In Table \ref{table:misclassification} and Table \ref{table:extended_result_cifar10}-\ref{table:extended_result_tim}, we present the results of the EDL models.
Interestingly, in Table \ref{table:misclassification}, we observe that EDL models often tend to produce lower AUROC and AUPR scores under distributional uncertainty measures for the misclassification detection task.
This property is desirable (see Section \ref{sec:proposed} and \ref{sec:toy_main}).
However, they achieve significantly lower OOD detection performance compared to the state-of-the-art competitive models (Table \ref{table:extended_result_cifar10}-\ref{table:extended_result_tim}).
Further, in Table \ref{table:rmsc}, we observe that the calibration performance of EDL models are also dropped than the other state-of-the-art OOD detection models.

\textbf{Existing DPN models \cite{dpn_nips_2018,dpn2_nips_2019}: }
DPN$_{fwd}$ and DPN$_{rev}$ models are trained only using the ADAM optimizer for all classification tasks \cite{adam_2014}. 
We could not use the SGD optimizer to train these models due to the complex RKL loss.
In contrast, we have not encountered such a problem for other models. 

For example, we use SGD to train the other models for C10 classification task. 
We observe that both DPN$_{fwd}$ and DPN$_{rev}$ models achieve lower classification accuracy than the other classifiers for this task (Table \ref{table:misclassification}).
For C100 and TIM classification tasks, we choose the ADAM optimizer for all models.  
We find that all the OOD detection models achieve similar classification accuracy for these two tasks (Table \ref{table:misclassification}).

Further, note that, the choice of larger values for the hyper-parameter, $\beta$ for the RKL in Eqn. \ref{eq_rev_kl_2} makes it difficult to optimize the network.
For our experiments, we choose the same set of hyper-parameters for DPN$_{fwd}$ and DPN$_{rev}$ models as suggested in the original paper \cite{dpn2_nips_2019}.
The concentration parameters for in-domain training examples are set to $100$ for the correct class and $1$ for the incorrect classes.
For OOD training examples, we choose the concentration parameters as $1$ for all classes.

\subsubsection{Description of the OOD Test Datasets}
\label{app:ood_description}
We use a wide-range of OOD datasets for our experiments, as described in the following.
For C10 and C100 classifiers, these input test images are resized to $32\times 32$, while for TIM classifiers, we resize them to $64\times 64$. 

\textit{TinyImageNet (Tiny)} \cite{db_tiny}. 
This dataset is used as an OOD test dataset \textit{only} for C-10 and C-100 classifiers.
Note that, for TinyImageNet classifiers, this is the in-domain test set.

This is a subset of the Imagenet dataset. 
We use the validation set, that contains $10,000$ test images from $200$ different image classes for our evaluation during test time.

\textit{CIFAR-10 and CIFAR-100} \cite{db_cifar}. 
This dataset is used as the OOD test dataset \textit{only} for TIM classifiers.
We use the validation set, that contains $10,000$ test images from $10$ and $100$ image classes respectively.

\textit{LSUN} \cite{db_lsun}. The Large-scale Scene 
UNderstanding dataset (LSUN) contains images of $10$ different scene categories.
We use its validation set, containing $10,000$ images, as an unknown OOD test set. 

\textit{Places 365} \cite{db_places365}. The validation set of this dataset consists of $36500$ images of $365$ scene categories.

\textit{Textures} \cite{db_textures} contains $5640$ 
textural images in the wild belonging to $47$ categories.

\textit{STL-10} contains $8,000$ images of natural images from $10$ different classes \cite{db_stl10}.

\subsection{Results for Confidence Calibration}
\label{sec:rmsc_app}
Calibration error measures if the confidence estimates produced by the classifier for its predictions misrepresent the empirical performance \cite{calib_emnlp_2015,calib_icml_2017,oe_iclr_2019}.
A well-calibrated classifier should produce the confidence probabilities that matches with the empirical frequency of correctness.
For example, if a classifier predicts an event with 90\% probability, we would
like it to be corrected for 90\%  of the time.
However, several studies have demonstrated that DNN classifiers tend to produce over-confidence in their predictions.

\begin{wraptable}{r}{5cm}{
\vspace*{-1.3em}
\caption{ Root mean square (RMS) calibration error. Lower scores are better.}
\label{table:rmsc}
\resizebox{5.0cm}{!}{%
\centering
\begin{tabular}{l|c|c|c}
\hline
& C10 & C100 & TIM 
\\ \hline
Baseline & 16.2{\tiny$\pm$0.0} & 6.6{\tiny$\pm$0.3} & 5.2{\tiny$\pm$0.0} 
\\
MCDP & 15.7{\tiny$\pm$0.1} & 6.7{\tiny$\pm$0.0} & 5.3{\tiny$\pm$0.2} 
\\
DE & 16.1{\tiny$\pm$NA} & 6.8{\tiny$\pm$NA} & 6.2{\tiny$\pm$NA} 
\\
EDL & 14.9{\tiny$\pm$0.2} & 17.7{\tiny$\pm$0.6} & 12.4{\tiny$\pm$0.1}
\\
OE & 6.4{\tiny$\pm$0.4} & 3.8{\tiny$\pm$0.1} & 4.2{\tiny$\pm$0.1} 
\\
DPN$_{rev}$ & 9.2{\tiny$\pm$0.4} & 10.4{\tiny$\pm$0.1} & 7.2{\tiny$\pm$0.5} 
\\
DPN$^+$ & \textbf{6.3{\tiny$\pm$0.3}} & 4.3{\tiny$\pm$0.0} & 2.8{\tiny$\pm$0.3} 
\\
DPN$^-$ & 6.5{\tiny$\pm$0.2} & \textbf{3.5{\tiny$\pm$0.1}} & \textbf{2.7{\tiny$\pm$0.3}} 
\\ \hline
\end{tabular}%
}}
\end{wraptable}

Several measures have been proposed to compute the calibration error of a classifier \cite{calib_icml_2017,oe_iclr_2019}.
In this paper, we use the Root Mean Square (RMS) Calibration Error that computes the square root of the expected squared difference between confidence and accuracy at a confidence level \cite{oe_iclr_2019}.
Since the confidence values can be distributed non-uniformly, Hendrycks et al. \cite{oe_iclr_2019} proposed to partition the samples into multiple bins with dynamic ranges and measure the average confidence and accuracy of each bin to compute the calibration error.

A real-world classifier should provide calibrated probabilities on both in- and out-of-distribution examples.
Hence, Hendrycks et al. \cite{oe_iclr_2019} proposed to incorporate OOD test examples in the calculation of the RMS calibration error.
Since the OOD examples do not belong to any of the in-domain classes, these examples are always considered to be incorrectly classified.
Hence, the classifier should produce low confidence for these OOD test examples.

In Table \ref{table:rmsc} we present a comparative results for RMS calibration error \cite{oe_iclr_2019}.
We take an equal number of OOD test examples as the in-domain test samples for this experiment.
For C10 and C100 classifiers, we use $5,000$ in-domain test examples from and take $5,000$ OOD test examples from STL-10 dataset.
For TIM classifiers, we use $5,000$ in-domain test examples and $5,000$ OOD test examples from CIFAR-100.
We apply the soft-max temperature scaling to report the calibration error in Table \ref{table:rmsc}.
Note that, EDL models do not apply soft-max activation to produce their probability scores.
Hence, we use `temperature translating' where we instead add the temperature parameter to the logit outputs.
We can Table \ref{table:rmsc} that our proposed DPN$^-$ models achieve comparable performance with the OE and DPN$^+$ models for C10.
For C100 and TIM, our DPN$^-$ models outperform other comparative systems.

\section{Derivations of different measures}
\label{sec:derivations}

\subsection{Differential Entropy for a Dirichlet}
\label{sec:d_ent}
Differential Entropy of a Dirichlet distribution can be calculated as follows:
\begin{equation}\small
\begin{split}
\label{eq:diff_ent_derivation}
 \mathcal{H}[p({\bm \mu}| {\bm x^*}, D_{in})]
& = - \int p({\bm \mu}| {\bm x^*}, D_{in}) \ln p({\bm \mu}| {\bm x^*}, D_{in}) d{\bm \mu}\\
& = \sum_{c=1}^{K} \ln \Gamma(\alpha_c) - \ln\Gamma(\alpha_0)
- \sum_{c=1}^{K} (\alpha_c-1)(\psi(\alpha_c)-\psi(\alpha_0))
\end{split}
\end{equation}
where, $\alpha_c$ is a function of $x^*$. 
$\Gamma$ and $\psi$ denotes the Gamma and digamma functions respectively.

\subsection{Mutual Information of a Dirichlet }
The mutual information of the labels $y$ and the categorical $\bm \mu$ of a DPN is computed as:
\begin{equation}
\begin{split}
\mathcal{I}[y, \bm \mu| \bm x^*, \hat{\bm \theta}]
 &= \sum_{c=1}^{K} \frac{\alpha_c}{\alpha_0}\big[ \psi(\alpha_c+1)-\psi(\alpha_0+1) -\ln \frac{\alpha_c}{\alpha_0}  \big]
\end{split}
\end{equation}

\subsection{KL Divergence between two Gaussians}
The KL divergence from a Gaussian distribution $\mathcal{N}_1(\bm{\mu}_1, \Sigma_1)$ to Gaussian distribution, $\mathcal{N}_2(\bm{\mu}_2, \Sigma_2)$ is computed as follows:
\begin{equation}
\begin{split}
KL(~\mathcal{N}_2 ~~||~~ \mathcal{N}_1) 
:= \frac{1}{2}\Big[tr(\Sigma_1^{-1}\Sigma_2) ~~- d + \frac{det(\Sigma_1)}{det(\Sigma_0)}
 + (\bm{\mu_1 - \mu_2})^T \Sigma_1^{-1} (\bm{\mu_1 - \mu_2})  \Big]
\end{split}
\end{equation}
where, $d$ is the dimension of $\bm \mu_1$ or $\bm \mu_2$.
$det(\Sigma)$ represents the determinant of $\Sigma$.
$tr$ computes the trace of a matrix.

\section{Extended Results}
\label{sec:extended}
In the following, we present an extended version of the results for a wide range of OOD test datasets along with an additional uncertainty measure i.e, entropy.
We also report the results for the deep ensemble (DE) framework using an ensemble of $3$ models.

In Table \ref{table:misclassification}, we present the results for misclassification detection along with the classification accuracy for different approaches.
Note that the DPN models achieve higher AUROC and AUPR scores even for distributional uncertainty measures such as MI, precision ($\alpha_0$), and D.Ent.

In Table \ref{table:kl_div_app}, we present an extended version of Table \ref{table:gauss_div} along with additional OOD datasets for each classification tasks.
In Table \ref{table:extended_result_cifar10}-\ref{table:extended_result_tim}, we present the additional results for OOD detection performance.

\begin{table}[htbp]
\centering
\caption{ Classification accuracy and misclassified image detection. 
	Here, we report the  (mean $\pm$ standard deviation) of 3 runs for each framework.
	Note that AUPR may \textit{not} be an ideal metric for comparison, as it depends on the \textit{base rates} i.e the number of misclassified examples v.s correctly classified predictions.
	That is, AUPR scores are comparable when the models achieve similar classification accuracy.
	Our DPN$^-$ models achieve comparable performance foe misclassified image detection using the AUROC metric.
}
\label{table:misclassification}
\vspace{0.5em}
\caption*{ (a) C10 classification task}
\resizebox{14.0cm}{!}{%
\begin{tabular}{l|ccccc|ccccc|c}
\hline
\multicolumn{1}{c|}{} 
& \multicolumn{5}{c}{AUROC} & \multicolumn{5}{c}{AUPR} &  \multirow{2}{*}{Acc. }
\\
\multicolumn{1}{c|}{} 
& Max.P & Ent & MI & $\alpha_0$ & D.Ent 
& Max.P & Ent & MI & $\alpha_0$ & D.Ent \\ \cline{2-12}
Baseline 
& 93.3{\tiny$\pm$0.1}  & 93.4{\tiny$\pm$0.1}  &-  &-  &- 
& 43.9{\tiny$\pm$0.7}  & 47.0{\tiny$\pm$0.3}  &-  &-  &- & 94.1{\tiny $\pm$0.0}
\\
MCDP 
& \textbf{93.6{\tiny$\pm$0.2}}  & \textbf{93.6{\tiny$\pm$0.2}}  & 93.2{\tiny$\pm$0.1}  &- &- 
& 46.1{\tiny$\pm$2.0}  & 46.5{\tiny$\pm$1.9}  & 40.9{\tiny$\pm$1.5} &- &- & 94.2{\tiny $\pm$0.1}
\\
DE &
93.5{\tiny$\pm$NA} & 93.5{\tiny$\pm$NA} & 92.7{\tiny$\pm$NA} &- &-  
& 45.6{\tiny$\pm$NA} & 46.6{\tiny$\pm$NA} & 39.8{\tiny$\pm$NA} &- &- 
& 94.0 {\tiny$\pm$NA}
\\
EDL 
& 91.3{\tiny$\pm$0.0}  & 91.2{\tiny$\pm$0.0}  & 80.1{\tiny$\pm$0.5}  & 76.7{\tiny$\pm$0.7}  & 89.7{\tiny$\pm$0.0}  & 44.8{\tiny$\pm$0.2}  & 43.8{\tiny$\pm$0.1}  & 21.7{\tiny$\pm$0.5}  & 18.1{\tiny$\pm$0.5}  & 37.9{\tiny$\pm$0.1} &
93.1{\tiny $\pm$0.0}
\\
OE 
& 92.0{\tiny$\pm$0.0}  & 91.6{\tiny$\pm$0.0}  & - &-  &- 
& 35.3{\tiny$\pm$0.8}  & 33.6{\tiny$\pm$0.8}  &-  &-  &- & 94.2{\tiny $\pm$0.1}
\\
DPN$_{fwd}$ 
& 90.3{\tiny$\pm$0.2}  & 90.1{\tiny$\pm$0.2}  & 88.6{\tiny$\pm$0.2}  & 88.0{\tiny$\pm$0.2}  & 88.0{\tiny$\pm$0.1}  
& 49.2{\tiny$\pm$0.9}  & 47.7{\tiny$\pm$0.8}  & 43.3{\tiny$\pm$0.4}  & 41.6{\tiny$\pm$0.4}  & 40.8{\tiny$\pm$0.6} & 88.3{\tiny $\pm$ 0.2}
\\
DPN$_{rev}$ 
& 89.6{\tiny$\pm$0.1}  & 89.4{\tiny$\pm$0.1}  & 88.7{\tiny$\pm$0.2}  & 88.7{\tiny$\pm$0.2}  & 89.0{\tiny$\pm$0.2}  
& \textbf{50.0{\tiny$\pm$0.8}}  & 48.8{\tiny$\pm$0.7}  & 46.1{\tiny$\pm$0.9}  & 45.8{\tiny$\pm$0.8}  & 47.7{\tiny$\pm$0.7} & 90.6{\tiny $\pm$0.0}
\\
DPN$^+$  
& 92.2{\tiny$\pm$0.3}  & 91.7{\tiny$\pm$0.3}  & 90.3{\tiny$\pm$0.1}  & 90.3{\tiny$\pm$0.1}  & 90.5{\tiny$\pm$0.2}  
& 36.6{\tiny$\pm$0.5}  & 34.9{\tiny$\pm$0.7}  & 31.2{\tiny$\pm$0.8}  & 31.2{\tiny$\pm$0.8}  & 31.6{\tiny$\pm$0.7} & 94.0{\tiny $\pm$ 0.1}
\\
DPN$^-$  
& 92.6{\tiny$\pm$0.1}  & 92.2{\tiny$\pm$0.1}  & 89.9{\tiny$\pm$0.0}  & 89.9{\tiny$\pm$0.0}  & 66.2{\tiny$\pm$0.7}  
& 37.2{\tiny$\pm$0.7}  & 35.1{\tiny$\pm$0.6}  & 31.3{\tiny$\pm$0.4}  & 30.6{\tiny$\pm$0.4}  & 17.1{\tiny$\pm$0.4} & \textbf{94.4{\tiny $\pm$0.0}}
\\ \hline
\end{tabular}
}
\bigskip
\caption*{ (b) C100 classification task}
\resizebox{14.0cm}{!}{%
\begin{tabular}{l|ccccc|ccccc|c}
\hline
\multicolumn{1}{c|}{} 
& \multicolumn{5}{c}{AUROC} & \multicolumn{5}{c}{AUPR} &  \multirow{2}{*}{Acc. }\\
& Max.P & Ent & MI & $\alpha_0$ & D.Ent 
& Max.P & Ent & MI & $\alpha_0$ & D.Ent \\ \cline{2-12}
Baseline 
& 86.8{\tiny$\pm$0.1}  & 87.0{\tiny$\pm$0.1}  & - & -  & -  
& 68.4{\tiny$\pm$0.4}  & 69.2{\tiny$\pm$0.3}  & - & - & - & 72.3{\tiny $\pm$0.0}
\\
MCDP 
& 87.2{\tiny$\pm$0.0}  & \textbf{87.3{\tiny$\pm$0.0}}  & 83.3{\tiny$\pm$0.3} &- & - 
& 69.1{\tiny$\pm$0.3}  & 69.3{\tiny$\pm$0.3}  & 53.9{\tiny$\pm$0.5} & - & - 
& \textbf{72.7{\tiny $\pm$0.1}}
\\
DE
& 87.0{\tiny $\pm$NA} & 87.1{\tiny $\pm$NA} & 83.4{\tiny $\pm$NA} &- &- 
& 69.2{\tiny $\pm$NA} & \textbf{69.7{\tiny $\pm$NA}} & 56.2{\tiny $\pm$NA} &- &-
& 72.2{\tiny $\pm$NA}
\\
EDL 
& 85.8{\tiny$\pm$0.3}  & 85.0{\tiny$\pm$0.3}  & 44.4{\tiny$\pm$1.0}  & 43.4{\tiny$\pm$1.1}  & 55.7{\tiny$\pm$0.7}  & 69.3{\tiny$\pm$1.1}  & 68.5{\tiny$\pm$1.0}  & 28.5{\tiny$\pm$1.0}  & 28.0{\tiny$\pm$1.0}  & 36.3{\tiny$\pm$1.4}
& 70.4{\tiny $\pm$0.3}
\\
OE 
& 86.9{\tiny$\pm$0.0}  & 86.9{\tiny$\pm$0.1}  & -  & -  & - 
& 67.7{\tiny$\pm$0.3}  & 66.9{\tiny$\pm$0.4}  & - & - & - & 71.6{\tiny $\pm$0.0}		
\\
DPN$_{rev}$ 
& 79.3{\tiny$\pm$0.1}  & 78.5{\tiny$\pm$0.1}  & 73.5{\tiny$\pm$0.1}  & 73.1{\tiny$\pm$0.1}  & 75.7{\tiny$\pm$0.1}  
& 65.3{\tiny$\pm$0.4}  & 64.1{\tiny$\pm$0.3}  & 58.4{\tiny$\pm$0.3}  & 57.9{\tiny$\pm$0.3}  & 61.2{\tiny$\pm$0.3} & 71.1{\tiny $\pm$0.1}
\\
DPN$^+$  
& 86.5{\tiny$\pm$0.1}  & 86.5{\tiny$\pm$0.1}  & 81.2{\tiny$\pm$0.0}  & 81.3{\tiny$\pm$0.0}  & 81.9{\tiny$\pm$0.1}  
& 66.8{\tiny$\pm$0.3}  & 66.3{\tiny$\pm$0.3}  & 57.8{\tiny$\pm$0.2}  & 57.8{\tiny$\pm$0.2}  & 59.2{\tiny$\pm$0.3} & 72.1{\tiny $\pm$0.1}
\\
DPN$^-$  
& 86.4{\tiny$\pm$0.1}  & 86.5{\tiny$\pm$0.1}  & 82.3{\tiny$\pm$0.0}  & 82.3{\tiny$\pm$0.0}  & 81.7{\tiny$\pm$0.1}
& 67.0{\tiny$\pm$0.5}  & 66.6{\tiny$\pm$0.3}  & 58.9{\tiny$\pm$0.2}  & 58.9{\tiny$\pm$0.2}  & 59.1{\tiny$\pm$0.2} & 72.3{\tiny $\pm$0.1}
\\ \hline
\end{tabular}
}
\bigskip
\caption*{ (c) TIM classification task}
\resizebox{14.0cm}{!}{%
\begin{tabular}{l|ccccc|ccccc|c}
\hline
\multicolumn{1}{c|}{} 
& \multicolumn{5}{c}{AUROC} & \multicolumn{5}{c}{AUPR} & \multirow{2}{*}{Acc. }
\\
\multicolumn{1}{c|}{} 
& Max.P & Ent & MI & $\alpha_0$ & D.Ent 
& Max.P & Ent & MI & $\alpha_0$ & D.Ent \\ \cline{2-12}
Baseline 
& 86.7{\tiny$\pm$0.0}  & 86.8{\tiny$\pm$0.1}  &-  &-  &- 
& 77.2{\tiny$\pm$0.1}  & 77.1{\tiny$\pm$0.3}  &-  &- & - & 62.5{\tiny $\pm$0.2}
\\
MCDP 
& 86.6{\tiny$\pm$0.1}  & 86.4{\tiny$\pm$0.1}  & 83.3{\tiny$\pm$0.3} &- &- 
& 76.8{\tiny$\pm$0.3}  & 76.4{\tiny$\pm$0.3}  & 67.2{\tiny$\pm$1.2} &- &- 
& \textbf{62.7{\tiny $\pm$0.2}}
\\
DE 
& \textbf{86.8{\tiny$\pm$NA}} & \textbf{86.8{\tiny$\pm$NA}} & 83.3{\tiny$\pm$NA} &- &- 
& 77.2{\tiny$\pm$NA} & 77.0{\tiny$\pm$NA} & 67.6{\tiny$\pm$NA} &- &- & 62.6{\tiny$\pm$NA}
\\
EDL
& 85.9{\tiny$\pm$0.2}  & 83.6{\tiny$\pm$0.1}  & 73.0{\tiny$\pm$0.6}  & 72.7{\tiny$\pm$0.6}  & 75.5{\tiny$\pm$0.4}  & 77.0{\tiny$\pm$0.4}  & 73.2{\tiny$\pm$0.4}  & 62.1{\tiny$\pm$0.5}  & 61.9{\tiny$\pm$0.6}  & 64.5{\tiny$\pm$0.4}
& 60.9{\tiny $\pm$0.1}
\\
OE 
& 85.9{\tiny$\pm$0.2}  & 85.8{\tiny$\pm$0.1}  & -  & - & -
& \textbf{77.7{\tiny$\pm$0.4}}  & 77.3{\tiny$\pm$0.2}  & -  & - & - & 59.8{\tiny $\pm$0.2}
\\
DPN$_{rev}$  
& 81.9{\tiny$\pm$0.3}  & 81.0{\tiny$\pm$0.2}  & 72.2{\tiny$\pm$0.7}  & 70.2{\tiny$\pm$0.9}  & 78.3{\tiny$\pm$0.3}  
& 75.0{\tiny$\pm$0.3}  & 73.4{\tiny$\pm$0.4}  & 61.4{\tiny$\pm$0.9}  & 59.2{\tiny$\pm$0.9}  & 70.0{\tiny$\pm$0.5} & 60.5{\tiny $\pm$0.2}
\\
DPN$^+$   
& 85.7{\tiny$\pm$0.2}  & 85.7{\tiny$\pm$0.1}  & 78.3{\tiny$\pm$0.4}  & 78.7{\tiny$\pm$0.5}  & 79.7{\tiny$\pm$0.2}  
& 77.4{\tiny$\pm$0.4}  & 76.7{\tiny$\pm$0.2}  & 66.3{\tiny$\pm$0.5}  & 66.4{\tiny$\pm$0.5}  & 68.5{\tiny$\pm$0.4} & 59.7{\tiny $\pm$0.1}
\\
DPN$^-$  
& 85.4{\tiny$\pm$0.1}  & 85.0{\tiny$\pm$0.0}  & 79.1{\tiny$\pm$0.5}  & 79.4{\tiny$\pm$0.4}  & 79.9{\tiny$\pm$0.2}  
& 76.9{\tiny$\pm$0.1}  & 76.2{\tiny$\pm$0.1}  & 67.3{\tiny$\pm$0.7}  & 67.4{\tiny$\pm$0.7}  & 69.4{\tiny$\pm$0.4} & 59.4{\tiny $\pm$0.1}
\\\hline
\end{tabular}
}
\end{table}

\begin{table}[htbp]
\centering
\caption{ KL-divergence scores from the distribution of uncertainty values of missclassified and correctly predicted examples to the OOD examples.
	Higher scores are desirable as it indicates greater gap between in-domain and OOD examples.
	We report the  (mean $\pm$ standard deviation) of 3 runs for each frameworks.}
\label{table:kl_div_app}
\vspace{0.5em}
\caption*{ (a) C10 classification task}
\vspace{0.5em}
\resizebox{12.0cm}{!}{%
\begin{tabular}{l|cc|cc|cc|cc|cc}
\hline
\multicolumn{1}{c|}{OOD} & 
\multicolumn{2}{c|}{STL-10} & \multicolumn{2}{c|}{Tiny} & \multicolumn{2}{c|}{LSUN} & 
\multicolumn{2}{c|}{Places365} & \multicolumn{2}{c}{Textures}
\\ \hline
\multicolumn{1}{c|}{} & 
Miss & Correct & Miss & Correct & Miss & Correct & 
Miss & Correct & Miss & Correct \\

DPN$_{fwd}$ 
 & \textbf{1.6{\tiny$\pm$0.1}}  & 1.6{\tiny$\pm$0.4} 
 & 1.1{\tiny$\pm$0.2}  & 6.1{\tiny$\pm$0.8} 
 & 1.4{\tiny$\pm$0.2}  & 6.7{\tiny$\pm$0.7} 
 & 1.3{\tiny$\pm$0.2}  & 6.6{\tiny$\pm$0.7} 
 & 2.8{\tiny$\pm$0.4}  & 12.7{\tiny$\pm$3.0} \\

DPN$_{rev}$ 
 & 0.1{\tiny$\pm$0.0}  & 5.7{\tiny$\pm$0.7} 
 & 1.4{\tiny$\pm$0.2}  & 12.6{\tiny$\pm$0.8} 
 & 2.1{\tiny$\pm$0.3}  & 13.9{\tiny$\pm$1.0} 
 & 1.6{\tiny$\pm$0.2}  & 13.2{\tiny$\pm$0.8} 
 & 3.5{\tiny$\pm$0.1}  & 16.3{\tiny$\pm$0.6} \\ 

DPN$^+$ 
 & 0.3{\tiny$\pm$0.0}  & 4.7{\tiny$\pm$0.5} 
 & 1.5{\tiny$\pm$0.2}  & 12.1{\tiny$\pm$1.9} 
 & 2.0{\tiny$\pm$0.2}  & 12.7{\tiny$\pm$2.0} 
 & 1.5{\tiny$\pm$0.2}  & 12.2{\tiny$\pm$1.9} 
 & \textbf{3.7{\tiny$\pm$0.1}}  & 15.4{\tiny$\pm$1.9} \\

DPN$^-$
 & 0.5{\tiny$\pm$0.0}  & \textbf{10.6{\tiny$\pm$0.8}}
 & \textbf{2.1{\tiny$\pm$0.1}}  & \textbf{20.7{\tiny$\pm$1.2}} 
 & \textbf{2.4{\tiny$\pm$0.1}}  & \textbf{22.5{\tiny$\pm$1.3}} 
 & \textbf{2.1{\tiny$\pm$0.1}}  & \textbf{20.9{\tiny$\pm$1.2} }
 & 2.9{\tiny$\pm$0.1}  & \textbf{20.4{\tiny$\pm$1.2}}  \\

\hline
\end{tabular}
}
\bigskip
\caption*{ (a) C100 classification task}
\vspace{0.5em}
\resizebox{13.5cm}{!}{%
\begin{tabular}{l|cc|cc|cc|cc|cc}
\hline
\multicolumn{1}{c|}{OOD} & 
\multicolumn{2}{c|}{STL-10} & \multicolumn{2}{c|}{Tiny} & \multicolumn{2}{c|}{LSUN} & 
\multicolumn{2}{c|}{Places365} & \multicolumn{2}{c}{Textures}
\\ \hline
\multicolumn{1}{c|}{} & 
Miss & Correct & Miss & Correct & Miss & Correct & 
Miss & Correct & Miss & Correct \\

DPN$_{rev}$ 
 & 2.8{\tiny$\pm$0.1}  & 9.4{\tiny$\pm$0.1} 
 & 1.8{\tiny$\pm$0.0}  & 6.6{\tiny$\pm$0.1} 
 & 2.6{\tiny$\pm$0.0}  & 8.6{\tiny$\pm$0.1} 
 & 2.1{\tiny$\pm$0.0}  & 7.3{\tiny$\pm$0.1} 
 & 1.1{\tiny$\pm$0.0}  & 4.0{\tiny$\pm$0.1}   \\ 

DPN$^+$ 
  & 2.7{\tiny$\pm$0.0}  & 9.3{\tiny$\pm$0.3} 
 & 1.8{\tiny$\pm$0.0}  & 7.2{\tiny$\pm$0.2} 
 & 2.6{\tiny$\pm$0.0}  & 9.1{\tiny$\pm$0.2} 
 & 2.1{\tiny$\pm$0.0}  & 7.8{\tiny$\pm$0.2} 
 & 0.5{\tiny$\pm$0.0}  & 3.5{\tiny$\pm$0.1} \\ 

DPN$^-$
 & \textbf{61.2{\tiny$\pm$3.6}}  & \textbf{330.1{\tiny$\pm$34.2} }
 & \textbf{44.5{\tiny$\pm$2.8}}  & \textbf{244.2{\tiny$\pm$25.2} }
 & \textbf{49.9{\tiny$\pm$3.5}}  & \textbf{272.0{\tiny$\pm$29.3}} 
 & \textbf{44.1{\tiny$\pm$3.0}}  & \textbf{241.7{\tiny$\pm$24.8}} 
 & \textbf{18.1{\tiny$\pm$1.3}}  & \textbf{105.0{\tiny$\pm$10.2}}  \\
\hline
\end{tabular}
}
\bigskip
\caption*{ (a) TIM classification task}
\vspace{0.5em}
\resizebox{9.5cm}{!}{%
\begin{tabular}{l|cc|cc|cc}
\hline
\multicolumn{1}{c|}{OOD} & 
\multicolumn{2}{c|}{STL-10} & \multicolumn{2}{c|}{CIFAR-10} & \multicolumn{2}{c}{CIFAR-100} 
\\ \hline
\multicolumn{1}{c|}{} & 
Miss & Correct & Miss & Correct & Miss & Correct \\
DPN$_{rev}$ 
 & 9.0{\tiny$\pm$0.6}  & 9.2{\tiny$\pm$1.0} 
 & 2.3{\tiny$\pm$0.4}  & 3.2{\tiny$\pm$0.5} 
 & 2.3{\tiny$\pm$0.4}  & 3.1{\tiny$\pm$0.5} 
\\ 

DPN$^+$ 
 & 21.1{\tiny$\pm$3.4}  & 27.1{\tiny$\pm$3.9} 
 & 17.7{\tiny$\pm$3.0}  & 23.5{\tiny$\pm$3.5} 
 & 17.3{\tiny$\pm$3.0}  & 23.1{\tiny$\pm$3.5} 
\\

DPN$^-$
 & \textbf{729.8{\tiny$\pm$12.4}}  & \textbf{1360.4{\tiny$\pm$94.5}} 
 & \textbf{664.9{\tiny$\pm$11.6}}  & \textbf{1241.4{\tiny$\pm$79.8}} 
 & \textbf{606.9{\tiny$\pm$11.4}}  & \textbf{1134.5{\tiny$\pm$72.1}} 
\\ \hline
\multicolumn{7}{c}{} \\

\end{tabular}
}
\resizebox{9.5cm}{!}{%
	\begin{tabular}{l|cc|cc|cc}
		\hline
		\multicolumn{1}{c|}{OOD} & 
		\multicolumn{2}{c|}{LSUN} & \multicolumn{2}{c|}{Places365} & \multicolumn{2}{c}{Textures} 
		\\ \hline
		\multicolumn{1}{c|}{} & 
		Miss & Correct & Miss & Correct & Miss & Correct \\
		DPN$_{rev}$ 
		& 9.7{\tiny$\pm$0.6}  & 9.9{\tiny$\pm$1.0} 
		& 8.3{\tiny$\pm$0.7}  & 8.5{\tiny$\pm$1.0} 
		& 4.4{\tiny$\pm$0.4}  & 4.8{\tiny$\pm$0.7} \\ 
		
		DPN$^+$ 
		& 21.5{\tiny$\pm$3.4}  & 27.5{\tiny$\pm$3.9} 
		& 20.6{\tiny$\pm$3.3}  & 26.5{\tiny$\pm$3.8} 
		& 14.7{\tiny$\pm$2.6}  & 20.0{\tiny$\pm$3.0} \\
		
		DPN$^-$
		& \textbf{731.6{\tiny$\pm$12.0}}  & \textbf{1363.4{\tiny$\pm$91.5}} 
		& \textbf{713.2{\tiny$\pm$11.7}}  & \textbf{1330.5{\tiny$\pm$89.2}} 
		& \textbf{465.7{\tiny$\pm$8.0} } & \textbf{873.2{\tiny$\pm$62.7}}  \\
		\hline
	\end{tabular}
}
\end{table}

\begin{table*}[!htbp]
\centering
\caption{ Results of OOD detection for C10.
	We report (mean $\pm$ standard deviation) of three different models. Description of these OOD datasets are provided in Appendix \ref{app:ood_description}. }
\label{table:extended_result_cifar10}
\vspace*{0.5em}
\resizebox{14cm}{!}{%
\begin{tabular}{ll|ccccc|ccccc}
\hline
\multirow{2}{*}{} & \multirow{2}{*}{Methods} & \multicolumn{5}{c}{AUROC} & \multicolumn{5}{c}{AUPR} \\
\multicolumn{2}{c|}{} & Max.P & Ent. & MI & $\alpha_0$ & D-Ent & Max.P & Ent. & MI & $\alpha_0$& D-Ent \\ \hline
\\
\multirow{7}{*}{\rotatebox{90}{Tiny} }
& Baseline 
& 88.9{\tiny$\pm$0.0}  & 89.5{\tiny$\pm$0.0}  & -  & -  & -  & 85.0{\tiny$\pm$0.1}  & 86.7{\tiny$\pm$0.1}  & -  & -  & -  \\
& MCDP 
& 88.7{\tiny$\pm$0.1}  & 88.9{\tiny$\pm$0.0}  & 88.1{\tiny$\pm$0.1}  &- &- & 85.2{\tiny$\pm$0.1}  & 85.9{\tiny$\pm$0.1}  & 84.0{\tiny$\pm$0.1} &- &- \\
& DE 
& 88.9{\tiny$\pm$NA} & 89.0{\tiny$\pm$NA} & 87.8{\tiny$\pm$NA} &- &-  & 85.0{\tiny$\pm$NA} & 85.5{\tiny$\pm$NA} & 83.2{\tiny$\pm$NA} & - &- \\
& EDL
& 87.6{\tiny$\pm$0.1}  & 88.4{\tiny$\pm$0.2}  & 89.1{\tiny$\pm$0.3}  & 87.6{\tiny$\pm$0.4}  & 89.9{\tiny$\pm$0.2}  & 84.9{\tiny$\pm$0.1}  & 86.6{\tiny$\pm$0.1}  & 88.9{\tiny$\pm$0.3}  & 87.4{\tiny$\pm$0.4}  & 89.0{\tiny$\pm$0.1} \\
& OE  
& 98.2{\tiny$\pm$0.1}  & 98.3{\tiny$\pm$0.1}  & -  & - & -  & 98.3{\tiny$\pm$0.2}  & 98.3{\tiny$\pm$0.2}  & -  & -  & -  \\
& DPN$_{fwd}$ 
& 92.8{\tiny$\pm$1.0}  & 93.0{\tiny$\pm$1.0}  & 73.3{\tiny$\pm$0.3}  & 71.3{\tiny$\pm$0.4}  & 93.7{\tiny$\pm$1.0}  & 92.6{\tiny$\pm$1.1}  & 92.9{\tiny$\pm$1.1}  & 59.8{\tiny$\pm$0.7}  & 58.1{\tiny$\pm$0.8}  & 93.4{\tiny$\pm$1.3}  \\ 
& DPN$_{rev}$ 
& 97.5{\tiny$\pm$0.5}  & 97.6{\tiny$\pm$0.5}  & 97.8{\tiny$\pm$0.4}  & 97.8{\tiny$\pm$0.4}  & 97.7{\tiny$\pm$0.4}  & 97.5{\tiny$\pm$0.5}  & 97.6{\tiny$\pm$0.4}  & 97.6{\tiny$\pm$0.3}  & 97.6{\tiny$\pm$0.3}  & 97.7{\tiny$\pm$0.4}  \\  
& DPN$^+$  
& 98.0{\tiny$\pm$0.2}  & 98.0{\tiny$\pm$0.2}  & 98.0{\tiny$\pm$0.2}  & 98.0{\tiny$\pm$0.2}  & 98.0{\tiny$\pm$0.2}  & 98.0{\tiny$\pm$0.3}  & 98.0{\tiny$\pm$0.3}  & 97.9{\tiny$\pm$0.3}  & 97.9{\tiny$\pm$0.3}  & 97.9{\tiny$\pm$0.3}  \\
& DPN$^-$  
& \textbf{99.0{\tiny$\pm$0.1}}  & \textbf{99.0{\tiny$\pm$0.1}}  & \textbf{99.0{\tiny$\pm$0.1} } & 97.7{\tiny$\pm$0.1}  & 6.0{\tiny$\pm$0.3}  & 99.0{\tiny$\pm$0.1}  & \textbf{99.1{\tiny$\pm$0.1}}  & 98.9{\tiny$\pm$0.1}  & 94.9{\tiny$\pm$0.1}  & 32.4{\tiny$\pm$0.1}  \\ \hline
\\
\multirow{6}{*}{\rotatebox{90}{STL-10} } 
& Baseline 
& 75.9{\tiny$\pm$0.0}  & 76.2{\tiny$\pm$0.0}  & - & -  & - & 68.5{\tiny$\pm$0.1}  & 69.8{\tiny$\pm$0.1}  & -  & - & - \\
& MCDP 
& 76.2{\tiny$\pm$0.0}  & 76.2{\tiny$\pm$0.0}  & 76.0{\tiny$\pm$0.0} &- &- & 69.3{\tiny$\pm$0.0}  & 69.7{\tiny$\pm$0.0}  & 69.4{\tiny$\pm$0.1} & - & - \\
& DE & 76.0{\tiny$\pm$NA} & 76.0{\tiny$\pm$NA} & 75.6{\tiny$\pm$NA} & - &- & 68.5{\tiny$\pm$NA} & 68.9{\tiny$\pm$NA} & 67.6{\tiny$\pm$NA} & - &-
\\
& EDL 
 & 72.4{\tiny$\pm$0.1}  & 72.7{\tiny$\pm$0.1}  & 71.7{\tiny$\pm$0.1}  & 70.6{\tiny$\pm$0.1}  & 73.3{\tiny$\pm$0.1}  & 66.8{\tiny$\pm$0.0}  & 68.0{\tiny$\pm$0.0}  & 68.2{\tiny$\pm$0.2}  & 66.7{\tiny$\pm$0.3}  & 69.4{\tiny$\pm$0.0}
\\
& OE   
& 81.4{\tiny$\pm$1.2}  & 81.5{\tiny$\pm$1.2}  & -  & -  & -  & 80.8{\tiny$\pm$1.1}  & 80.8{\tiny$\pm$1.0}  & -  & -  & -  \\
& DPN$_{fwd}$ 
& 71.5{\tiny$\pm$1.3}  & 71.6{\tiny$\pm$1.3}  & 65.7{\tiny$\pm$0.2}  & 64.9{\tiny$\pm$0.2}  & 72.0{\tiny$\pm$1.5}  & 68.2{\tiny$\pm$2.1}  & 68.6{\tiny$\pm$2.1}  & 53.3{\tiny$\pm$0.6}  & 52.5{\tiny$\pm$0.7}  & 68.6{\tiny$\pm$2.7}   \\
& DPN$_{rev}$  
& 81.6{\tiny$\pm$1.7}  & 81.7{\tiny$\pm$1.7}  & 82.2{\tiny$\pm$1.7}  & 82.2{\tiny$\pm$1.6}  & 81.9{\tiny$\pm$1.7}  & 81.9{\tiny$\pm$1.7}  & 82.0{\tiny$\pm$1.7}  & 82.5{\tiny$\pm$1.7}  & 82.5{\tiny$\pm$1.6}  & 82.2{\tiny$\pm$1.7}  \\ 
& DPN$^+$ 
& 81.6{\tiny$\pm$1.4}  & 81.7{\tiny$\pm$1.3}  & 81.8{\tiny$\pm$1.2}  & 81.8{\tiny$\pm$1.2}  & 81.8{\tiny$\pm$1.2}  & 80.9{\tiny$\pm$1.2}  & 81.0{\tiny$\pm$1.3}  & 81.0{\tiny$\pm$1.2}  & 81.0{\tiny$\pm$1.2}  & 81.0{\tiny$\pm$1.2} \\ 
& DPN$^-$ 
& 84.7{\tiny$\pm$0.4}  & 84.8{\tiny$\pm$0.5}  & \textbf{85.3{\tiny$\pm$0.5}}  & 84.9{\tiny$\pm$0.5}  & 34.6{\tiny$\pm$0.4}  & 84.7{\tiny$\pm$0.6}  & 84.9{\tiny$\pm$0.6}  & \textbf{85.2{\tiny$\pm$0.6}}  & 82.0{\tiny$\pm$0.6}  & 42.4{\tiny$\pm$0.2} \\ 
\hline
\\ 
\multirow{6}{*}{\rotatebox{90}{LSUN} }
& Baseline 
& 90.3{\tiny$\pm$0.0}  & 91.0{\tiny$\pm$0.0}  & -  & -  & - & 86.6{\tiny$\pm$0.1}  & 88.5{\tiny$\pm$0.1}  & - & - & - \\
& MCDP 
& 90.6{\tiny$\pm$0.0}  & 90.8{\tiny$\pm$0.0}  & 90.2{\tiny$\pm$0.0} &- &- & 87.5{\tiny$\pm$0.1}  & 88.2{\tiny$\pm$0.0}  & 86.6{\tiny$\pm$0.1}  & - & - \\
& DE &
90.3{\tiny$\pm$NA} & 90.4{\tiny$\pm$NA} & 89.7{\tiny$\pm$NA} &- &- & 86.5{\tiny$\pm$NA} & 87.1{\tiny$\pm$NA} & 85.6{\tiny$\pm$NA} & - &-
\\
& EDL 
& 90.3{\tiny$\pm$0.0}  & 91.2{\tiny$\pm$0.0}  & 93.5{\tiny$\pm$0.0}  & 92.6{\tiny$\pm$0.0}  & 93.0{\tiny$\pm$0.0}  & 87.8{\tiny$\pm$0.0}  & 89.7{\tiny$\pm$0.0}  & 93.4{\tiny$\pm$0.0}  & 92.3{\tiny$\pm$0.0}  & 92.4{\tiny$\pm$0.0} \\
& OE  
& 98.4{\tiny$\pm$0.3}  & 98.4{\tiny$\pm$0.3}  & -  & -  & -  & 98.2{\tiny$\pm$0.4}  & 98.2{\tiny$\pm$0.4}  & -  & - & - \\
& DPN$_{fwd}$ 
& 93.5{\tiny$\pm$0.7}  & 93.7{\tiny$\pm$0.8}  & 72.6{\tiny$\pm$0.1}  & 70.6{\tiny$\pm$0.2}  & 94.9{\tiny$\pm$0.8}  & 93.2{\tiny$\pm$0.9}  & 93.5{\tiny$\pm$0.9}  & 58.6{\tiny$\pm$0.3}  & 57.0{\tiny$\pm$0.3}  & 94.4{\tiny$\pm$1.0}  \\ 
& DPN$_{rev}$ 
& 98.5{\tiny$\pm$0.4}  & 98.6{\tiny$\pm$0.3}  & 98.7{\tiny$\pm$0.3}  & 98.7{\tiny$\pm$0.3}  & 98.7{\tiny$\pm$0.3}  & 98.3{\tiny$\pm$0.4}  & 98.4{\tiny$\pm$0.3}  & 98.5{\tiny$\pm$0.3}  & 98.5{\tiny$\pm$0.3}  & 98.5{\tiny$\pm$0.3}   \\
& DPN$^+$ 
& 98.2{\tiny$\pm$0.3}  & 98.3{\tiny$\pm$0.4}  & 98.3{\tiny$\pm$0.4}  & 98.3{\tiny$\pm$0.4}  & 98.3{\tiny$\pm$0.4}  & 98.0{\tiny$\pm$0.5}  & 98.1{\tiny$\pm$0.5}  & 98.0{\tiny$\pm$0.4}  & 98.0{\tiny$\pm$0.4}  & 98.0{\tiny$\pm$0.4} \\
& DPN$^-$ 
& 99.2{\tiny$\pm$0.1}  & 99.2{\tiny$\pm$0.1}  & \textbf{99.3{\tiny$\pm$0.0}}  & 98.1{\tiny$\pm$0.1}  & 5.0{\tiny$\pm$0.2}  & 99.1{\tiny$\pm$0.1}  & \textbf{99.2{\tiny$\pm$0.1}}  & 99.1{\tiny$\pm$0.1}  & 95.9{\tiny$\pm$0.0}  & 32.1{\tiny$\pm$0.2} \\
\hline
\\
\multirow{6}{*}{\rotatebox{90}{Places365} }
& Baseline 
& 89.4{\tiny$\pm$0.0}  & 90.0{\tiny$\pm$0.0}  & -  & -  & - & 95.5{\tiny$\pm$0.0}  & 96.1{\tiny$\pm$0.0}  & - & -  & - \\ 
& MCDP 
& 89.5{\tiny$\pm$0.0}  & 89.7{\tiny$\pm$0.0}  & 89.0{\tiny$\pm$0.0} &- &- & 95.7{\tiny$\pm$0.0}  & 96.0{\tiny$\pm$0.0}  & 95.3{\tiny$\pm$0.0}   & - & - \\
& DE & 89.4{\tiny$\pm$NA} & 89.5{\tiny$\pm$NA} & 88.7{\tiny$\pm$NA} &- &- & 95.5{\tiny$\pm$NA} & 95.7{\tiny$\pm$NA} & 95.2{\tiny$\pm$NA} &- &- 
\\
& EDL 
& 88.5{\tiny$\pm$0.0}  & 89.4{\tiny$\pm$0.0}  & 91.3{\tiny$\pm$0.0}  & 90.3{\tiny$\pm$0.1}  & 91.2{\tiny$\pm$0.0}  & 95.6{\tiny$\pm$0.0}  & 96.2{\tiny$\pm$0.0}  & 97.3{\tiny$\pm$0.0}  & 96.8{\tiny$\pm$0.0}  & 97.1{\tiny$\pm$0.0}
\\
& OE 
& 98.1{\tiny$\pm$0.1}  & 98.2{\tiny$\pm$0.2}  & -  & - & -  & 99.4{\tiny$\pm$0.1}  & 99.4{\tiny$\pm$0.1}  & - & - & - \\ 
& DPN$_{fwd}$ 
& 93.3{\tiny$\pm$0.9}  & 93.5{\tiny$\pm$0.8}  & 72.1{\tiny$\pm$0.2}  & 70.1{\tiny$\pm$0.3}  & 94.4{\tiny$\pm$0.9}  & 97.9{\tiny$\pm$0.3}  & 98.0{\tiny$\pm$0.3}  & 82.4{\tiny$\pm$0.6}  & 81.4{\tiny$\pm$0.6}  & 98.2{\tiny$\pm$0.3}  \\
& DPN$_{rev}$ 
& 97.8{\tiny$\pm$0.4}  & 97.9{\tiny$\pm$0.4}  & 98.0{\tiny$\pm$0.4}  & 98.0{\tiny$\pm$0.4}  & 98.0{\tiny$\pm$0.4}  & 99.3{\tiny$\pm$0.1}  & 99.3{\tiny$\pm$0.1}  & 99.4{\tiny$\pm$0.1}  & 99.4{\tiny$\pm$0.1}  & 99.4{\tiny$\pm$0.1}  \\ 
& DPN$^+$ 
& 98.0{\tiny$\pm$0.3}  & 98.0{\tiny$\pm$0.3}  & 98.0{\tiny$\pm$0.3}  & 98.0{\tiny$\pm$0.3}  & 98.0{\tiny$\pm$0.3}  & 99.4{\tiny$\pm$0.1}  & 99.4{\tiny$\pm$0.1}  & 99.4{\tiny$\pm$0.1}  & 99.4{\tiny$\pm$0.1}  & 99.4{\tiny$\pm$0.1} \\
& DPN$^-$ 
& \textbf{98.9{\tiny$\pm$0.1}}  & \textbf{98.9{\tiny$\pm$0.1}}  & \textbf{98.9{\tiny$\pm$0.1}}  & 97.7{\tiny$\pm$0.1}  & 6.2{\tiny$\pm$0.4}  & \textbf{99.7{\tiny$\pm$0.0}}  & \textbf{99.7{\tiny$\pm$0.0}}  & \textbf{99.7{\tiny$\pm$0.0}}  & 98.6{\tiny$\pm$0.0}  & 61.0{\tiny$\pm$0.2} \\

\hline
\\
\multirow{6}{*}{\rotatebox{90}{Textures} }
& Baseline 
& 88.8{\tiny$\pm$0.0}  & 89.2{\tiny$\pm$0.0}  & -  & -  & - & 74.9{\tiny$\pm$0.1}  & 76.9{\tiny$\pm$0.2}  & -  & -  & - \\
& MCDP 
& 87.4{\tiny$\pm$0.1}  & 87.5{\tiny$\pm$0.1}  & 85.7{\tiny$\pm$0.2} &- &- & 73.3{\tiny$\pm$0.2}  & 74.2{\tiny$\pm$0.1}  & 66.9{\tiny$\pm$0.3} &- &-  
\\
& DE & 88.7{\tiny$\pm$NA} & 88.8{\tiny$\pm$NA} & 87.2{\tiny$\pm$NA} &- &-  & 74.8{\tiny$\pm$NA} & 75.4{\tiny$\pm$NA} & 72.1{\tiny$\pm$NA} &- &- 
\\
& EDL 
& 84.3{\tiny$\pm$0.4}  & 84.9{\tiny$\pm$0.5}  & 84.9{\tiny$\pm$0.9}  & 83.1{\tiny$\pm$1.0}  & 86.3{\tiny$\pm$0.6}  & 71.5{\tiny$\pm$0.4}  & 73.6{\tiny$\pm$0.5}  & 76.4{\tiny$\pm$1.1}  & 73.9{\tiny$\pm$1.3}  & 77.0{\tiny$\pm$0.6}  \\
& OE 
& 99.4{\tiny$\pm$0.0}  & 99.4{\tiny$\pm$0.1}  & - & -  & -  & 98.9{\tiny$\pm$0.2}  & 98.9{\tiny$\pm$0.3}  & - & - & - \\ 
& DPN$_{fwd}$ 
& 98.3{\tiny$\pm$0.4}  & 98.4{\tiny$\pm$0.4}  & 68.2{\tiny$\pm$0.2}  & 65.8{\tiny$\pm$0.1}  & 98.6{\tiny$\pm$0.2}  & 97.1{\tiny$\pm$0.6}  & 97.4{\tiny$\pm$0.6}  & 42.0{\tiny$\pm$0.2}  & 40.4{\tiny$\pm$0.2}  & 97.7{\tiny$\pm$0.6}  \\
& DPN$_{rev}$ 
& 99.4{\tiny$\pm$0.0}  & 99.3{\tiny$\pm$0.0}  & 99.4{\tiny$\pm$0.0}  & 99.4{\tiny$\pm$0.0}  & 99.4{\tiny$\pm$0.0}  & 98.5{\tiny$\pm$0.0}  & 98.3{\tiny$\pm$0.0}  & 98.5{\tiny$\pm$0.1}  & 98.4{\tiny$\pm$0.1}  & 98.4{\tiny$\pm$0.0}   \\ 
& DPN$^+$ 
& 99.5{\tiny$\pm$0.0}  & 99.6{\tiny$\pm$0.0}  & 99.5{\tiny$\pm$0.0}  & 99.5{\tiny$\pm$0.0}  & 99.5{\tiny$\pm$0.0}  & 99.1{\tiny$\pm$0.2}  & 99.1{\tiny$\pm$0.1}  & 99.1{\tiny$\pm$0.1}  & 99.1{\tiny$\pm$0.1}  & 99.1{\tiny$\pm$0.1}  \\
& DPN$^-$ 
& \textbf{99.7{\tiny$\pm$0.0}}  & \textbf{99.7{\tiny$\pm$0.0}}  & 99.5{\tiny$\pm$0.0}  & 97.8{\tiny$\pm$0.1}  & 3.6{\tiny$\pm$0.1}  & \textbf{99.4{\tiny$\pm$0.0}}  & \textbf{99.4{\tiny$\pm$0.1}}  & 98.8{\tiny$\pm$0.0}  & 90.1{\tiny$\pm$0.3}  & 21.3{\tiny$\pm$0.0} \\
 
\hline
\end{tabular}
}
\end{table*}

\begin{table*}[ht]
\centering
\caption{ Results of OOD image detection for C100.
	We report (mean $\pm$ standard deviation) of three different models. 
	Description of these OOD datasets are provided in Appendix \ref{app:ood_description}.}
\label{table:extended_result_cifar100}
\vspace*{0.5em}

\resizebox{14cm}{!}{%
\begin{tabular}{ll|ccccc|ccccc}
\hline
\multirow{2}{*}{} & \multirow{2}{*}{Methods} & \multicolumn{5}{c}{AUROC} & \multicolumn{5}{c}{AUPR} \\
& & Max.P & Ent. & MI & $\alpha_0$ & D-Ent & Max.P & Ent. & MI & $\alpha_0$& D-Ent \\ \hline

\multirow{6}{*}{\rotatebox{90}{Tiny} } 
\\
& Baseline 
& 68.8{\tiny$\pm$0.2}  & 71.4{\tiny$\pm$0.2}  & -  & -  & - & 66.6{\tiny$\pm$0.2}  & 70.2{\tiny$\pm$0.2}  & - & -  & -
\\
& MCDP 
& 69.7{\tiny$\pm$0.3}  & 70.2{\tiny$\pm$0.3}  & 70.6{\tiny$\pm$0.3} &- &- & 67.4{\tiny$\pm$0.3}  & 68.5{\tiny$\pm$0.2}  & 66.0{\tiny$\pm$0.2} &- &-
\\
& DE 
& 68.9{\tiny$\pm$NA} & 69.3{\tiny$\pm$NA} & 69.6{\tiny$\pm$NA} &- &- 
&66.7{\tiny$\pm$NA} & 67.7{\tiny$\pm$NA} & 66.3{\tiny$\pm$NA} & - &- 
\\
& EDL 
& 66.9{\tiny$\pm$0.2}  & 71.5{\tiny$\pm$0.2}  & 72.8{\tiny$\pm$0.5}  & 72.2{\tiny$\pm$0.6}  & 77.4{\tiny$\pm$0.1}  & 62.8{\tiny$\pm$0.4}  & 68.9{\tiny$\pm$0.6}  & 71.4{\tiny$\pm$0.7}  & 71.0{\tiny$\pm$0.7}  & 74.8{\tiny$\pm$0.4}
\\
& OE 
& 89.5{\tiny$\pm$1.0}  & 91.2{\tiny$\pm$0.9} & -  & -  & -  & 91.1{\tiny$\pm$0.9}  & 92.6{\tiny$\pm$0.8} & -  & -  & -
\\ 
& DPN$_{rev}$ 
& 81.2{\tiny$\pm$0.2}  & 82.4{\tiny$\pm$0.1}  & 83.8{\tiny$\pm$0.1}  & 83.8{\tiny$\pm$0.1}  & 83.5{\tiny$\pm$0.1}  & 84.7{\tiny$\pm$0.0}  & 86.1{\tiny$\pm$0.0}  & 87.6{\tiny$\pm$0.0}  & 87.6{\tiny$\pm$0.0}  & 87.1{\tiny$\pm$0.0}
\\

& DPN$^+$ 
& 85.9{\tiny$\pm$0.3}  & 88.1{\tiny$\pm$0.2}  & 92.2{\tiny$\pm$0.1}  & 92.2{\tiny$\pm$0.1}  & 92.3{\tiny$\pm$0.1}  & 88.0{\tiny$\pm$0.2}  & 90.0{\tiny$\pm$0.2}  & 92.7{\tiny$\pm$0.1}  & 92.7{\tiny$\pm$0.1}  & 92.8{\tiny$\pm$0.1}
\\

& DPN$^-$ 
& 89.2{\tiny$\pm$0.1}  & 90.7{\tiny$\pm$0.1}  & \textbf{94.5{\tiny$\pm$0.1}}  & \textbf{94.5{\tiny$\pm$0.1}}  & 38.1{\tiny$\pm$0.5}  & 91.4{\tiny$\pm$0.1}  & 92.7{\tiny$\pm$0.1}  & \textbf{95.1{\tiny$\pm$0.1}}  & \textbf{95.1{\tiny$\pm$0.1}}  & 55.7{\tiny$\pm$0.4}
\\ 
\hline
\\
 \multirow{6}{*}{\rotatebox{90}{STL-10} } 
& Baseline 
& 69.6{\tiny$\pm$0.0}  & 71.9{\tiny$\pm$0.0}  & - & - & - & 61.9{\tiny$\pm$0.1}  & 65.4{\tiny$\pm$0.1}  & - & - & -
\\
& MCDP 
& 70.7{\tiny$\pm$0.1}  & 71.2{\tiny$\pm$0.1}  & 71.6{\tiny$\pm$0.2} &- &- & 62.8{\tiny$\pm$0.1}  & 63.9{\tiny$\pm$0.2}  & 61.4{\tiny$\pm$0.1} &- &-
\\
& DE 
& 69.6{\tiny$\pm$NA} & 70.1{\tiny$\pm$NA} & 70.2{\tiny$\pm$NA} &- &-
& 62.0{\tiny$\pm$NA} & 63.0{\tiny$\pm$NA} & 60.9{\tiny$\pm$NA} & - &-
\\
& EDL  & 68.1{\tiny$\pm$0.2}  & 72.0{\tiny$\pm$0.2}  & 68.0{\tiny$\pm$0.6}  & 67.3{\tiny$\pm$0.7}  & 73.8{\tiny$\pm$0.4}  & 58.5{\tiny$\pm$0.6}  & 64.1{\tiny$\pm$0.4}  & 61.6{\tiny$\pm$1.0}  & 61.1{\tiny$\pm$1.1}  & 66.1{\tiny$\pm$0.7} 
\\
& OE  
& 91.2{\tiny$\pm$0.7}  & 92.7{\tiny$\pm$0.6} & - & - & - & 92.1{\tiny$\pm$0.6}  & 93.4{\tiny$\pm$0.5}   & - & - & -  
\\
& DPN$_{rev}$ 
& 87.2{\tiny$\pm$0.1}  & 88.1{\tiny$\pm$0.1}  & 89.3{\tiny$\pm$0.1}  & 89.3{\tiny$\pm$0.1}  & 89.0{\tiny$\pm$0.1}  & 88.5{\tiny$\pm$0.0}  & 89.6{\tiny$\pm$0.1}  & 91.0{\tiny$\pm$0.1}  & 91.1{\tiny$\pm$0.1}  & 90.5{\tiny$\pm$0.1} 
\\ 

& DPN$^+$  
& 89.1{\tiny$\pm$0.2}  & 90.8{\tiny$\pm$0.2}  & 95.0{\tiny$\pm$0.0}  & 95.0{\tiny$\pm$0.0}  & 94.8{\tiny$\pm$0.0}  & 90.0{\tiny$\pm$0.2}  & 91.7{\tiny$\pm$0.2}  & 94.7{\tiny$\pm$0.0}  & 94.7{\tiny$\pm$0.0}  & 94.6{\tiny$\pm$0.1} 
\\

& DPN$^-$ 
& 92.8{\tiny$\pm$0.1}  & 93.9{\tiny$\pm$0.1}  & \textbf{96.8{\tiny$\pm$0.1}}  & \textbf{96.8{\tiny$\pm$0.1}}  & 25.4{\tiny$\pm$0.4}  & 93.7{\tiny$\pm$0.1}  & 94.7{\tiny$\pm$0.1}  & \textbf{96.7{\tiny$\pm$0.1}}  & \textbf{96.7{\tiny$\pm$0.1}}  & 42.8{\tiny$\pm$0.3} 
\\  \hline
\\ 
\multirow{6}{*}{\rotatebox{90}{LSUN} } 
& Baseline 
& 72.5{\tiny$\pm$0.0}  & 75.0{\tiny$\pm$0.0}  & -  & - & - & 69.0{\tiny$\pm$0.1}  & 72.7{\tiny$\pm$0.1}  & - & -  & -
\\
& MCDP 
& 74.5{\tiny$\pm$0.1}  & 75.1{\tiny$\pm$0.1}  & 75.9{\tiny$\pm$0.2} &- &- & 70.8{\tiny$\pm$0.3}  & 71.9{\tiny$\pm$0.2}  & 70.4{\tiny$\pm$0.2}  &- &-
\\
& DE 
& 72.6{\tiny$\pm$NA} & 73.0{\tiny$\pm$NA} & 73.4{\tiny$\pm$NA} &- &- & 69.1{\tiny$\pm$NA} & 70.0{\tiny$\pm$NA} & 68.6{\tiny$\pm$NA} & - &- 
\\
& EDL & 67.6{\tiny$\pm$0.6}  & 72.3{\tiny$\pm$0.6}  & 72.8{\tiny$\pm$0.6}  & 72.3{\tiny$\pm$0.6}  & 76.7{\tiny$\pm$0.5}  & 62.3{\tiny$\pm$1.0}  & 69.3{\tiny$\pm$1.1}  & 72.9{\tiny$\pm$0.8}  & 72.5{\tiny$\pm$0.9}  & 76.0{\tiny$\pm$0.5}
\\
& OE 
& 92.2{\tiny$\pm$0.9}  & 93.7{\tiny$\pm$0.7}  & - & - & - & 93.7{\tiny$\pm$0.7}  & 94.9{\tiny$\pm$0.7}  & - & - & -
\\ 
& DPN$_{rev}$ 
& 86.7{\tiny$\pm$0.0}  & 87.9{\tiny$\pm$0.0}  & 89.3{\tiny$\pm$0.1}  & 89.3{\tiny$\pm$0.1}  & 88.9{\tiny$\pm$0.1}  & 89.2{\tiny$\pm$0.0}  & 90.5{\tiny$\pm$0.0}  & 92.0{\tiny$\pm$0.0}  & 92.0{\tiny$\pm$0.0}  & 91.5{\tiny$\pm$0.0} 
\\ 
& DPN$^+$ 
& 90.3{\tiny$\pm$0.3}  & 92.1{\tiny$\pm$0.3}  & 95.0{\tiny$\pm$0.1}  & 95.0{\tiny$\pm$0.1}  & 95.0{\tiny$\pm$0.1}  & 92.0{\tiny$\pm$0.2}  & 93.6{\tiny$\pm$0.2}  & 95.5{\tiny$\pm$0.1}  & 95.5{\tiny$\pm$0.1}  & 95.6{\tiny$\pm$0.1} 
\\
& DPN$^-$ 
& 92.8{\tiny$\pm$0.1}  & 94.0{\tiny$\pm$0.1}  & \textbf{96.5{\tiny$\pm$0.1}}  & \textbf{96.5{\tiny$\pm$0.1}}  & 31.5{\tiny$\pm$0.4}  & 94.3{\tiny$\pm$0.1}  & 95.3{\tiny$\pm$0.1}  & \textbf{97.0{\tiny$\pm$0.1}}  & 96.9{\tiny$\pm$0.1}  & 52.6{\tiny$\pm$0.3}
\\ 
\hline
\\
\multirow{6}{*}{\rotatebox{90}{Places365} } 
& Baseline 
& 70.7{\tiny$\pm$0.0}  & 73.2{\tiny$\pm$0.0}  & -  & - & - & 88.0{\tiny$\pm$0.0}  & 89.6{\tiny$\pm$0.0}  & - & -  & -
\\
& MCDP 
& 72.1{\tiny$\pm$0.1}  & 72.6{\tiny$\pm$0.1}  & 73.4{\tiny$\pm$0.2} &- &- & 88.6{\tiny$\pm$0.0}  & 89.0{\tiny$\pm$0.0}  & 88.4{\tiny$\pm$0.1} &- &-
\\
& DE 
&70.8{\tiny$\pm$NA} & 71.2{\tiny$\pm$NA} & 72.3{\tiny$\pm$NA} &- &- & 88.1{\tiny$\pm$NA} & 88.4{\tiny$\pm$NA} & 88.6{\tiny$\pm$NA} & - &-
\\
& EDL 
& 66.8{\tiny$\pm$0.5}  & 71.0{\tiny$\pm$0.6}  & 70.2{\tiny$\pm$0.3}  & 69.7{\tiny$\pm$0.3}  & 74.3{\tiny$\pm$0.4}  & 85.5{\tiny$\pm$0.5}  & 88.4{\tiny$\pm$0.4}  & 89.2{\tiny$\pm$0.2}  & 89.0{\tiny$\pm$0.3}  & 90.6{\tiny$\pm$0.1}
\\
& OE 
& 89.3{\tiny$\pm$1.0}  & 90.9{\tiny$\pm$0.9}  & - & - & - & 97.0{\tiny$\pm$0.3}  & 97.5{\tiny$\pm$0.3}  & - & - & -
\\ 
& DPN$_{rev}$ 
& 83.0{\tiny$\pm$0.1}  & 84.2{\tiny$\pm$0.1}  & 85.8{\tiny$\pm$0.0}  & 85.8{\tiny$\pm$0.1}  & 85.4{\tiny$\pm$0.0}  & 95.1{\tiny$\pm$0.0}  & 95.6{\tiny$\pm$0.0}  & 96.1{\tiny$\pm$0.0}  & 96.1{\tiny$\pm$0.0}  & 95.9{\tiny$\pm$0.0} 
\\ 

& DPN$^+$ 
& 87.1{\tiny$\pm$0.2}  & 89.3{\tiny$\pm$0.2}  & 92.9{\tiny$\pm$0.1}  & 92.9{\tiny$\pm$0.1}  & 93.0{\tiny$\pm$0.1}  & 96.3{\tiny$\pm$0.0}  & 96.9{\tiny$\pm$0.1}  & 97.9{\tiny$\pm$0.0}  & 97.9{\tiny$\pm$0.0}  & 97.9{\tiny$\pm$0.0} 
\\
& DPN$^-$  
& 89.9{\tiny$\pm$0.2}  & 91.3{\tiny$\pm$0.1}  & \textbf{94.5{\tiny$\pm$0.1}}  & \textbf{94.5{\tiny$\pm$0.1}}  & 37.9{\tiny$\pm$0.5}  & 97.2{\tiny$\pm$0.0}  & 97.6{\tiny$\pm$0.0}  & \textbf{98.4{\tiny$\pm$0.0}}  & \textbf{98.4{\tiny$\pm$0.0}}  & 79.8{\tiny$\pm$0.2}
\\ 
\hline
\\
\multirow{6}{*}{\rotatebox{90}{Textures} } 
& Baseline 
& 62.8{\tiny$\pm$0.2}  & 64.7{\tiny$\pm$0.2}  & - & - & -  & 43.8{\tiny$\pm$0.2}  & 45.9{\tiny$\pm$0.2}  & - & -  & -
\\
& MCDP 
& 64.3{\tiny$\pm$0.3}  & 64.6{\tiny$\pm$0.2}  & 67.2{\tiny$\pm$0.2} &- &- & 44.9{\tiny$\pm$0.3}  & 45.3{\tiny$\pm$0.2}  & 49.6{\tiny$\pm$0.0} &- &-
\\
& DE 
& 62.9{\tiny$\pm$NA} & 63.0{\tiny$\pm$NA} & 69.0{\tiny$\pm$NA} &- &-
& 43.9{\tiny$\pm$NA} & 44.1{\tiny$\pm$NA} & 56.1{\tiny$\pm$NA} &- &-
\\
& EDL 
& 66.9{\tiny$\pm$0.7}  & 72.1{\tiny$\pm$0.8}  & 73.7{\tiny$\pm$0.5}  & 73.1{\tiny$\pm$0.4}  & 78.1{\tiny$\pm$1.4}  & 47.5{\tiny$\pm$2.2}  & 55.5{\tiny$\pm$2.4}  & 60.1{\tiny$\pm$1.0}  & 59.5{\tiny$\pm$0.9}  & 64.8{\tiny$\pm$1.4} 
\\
& OE 
& 79.7{\tiny$\pm$1.0}  & 81.2{\tiny$\pm$1.0}  & -  & -  & - & 71.7{\tiny$\pm$1.5}  & 73.2{\tiny$\pm$1.6}  & - & - & -
\\
& DPN$_{rev}$ 
& 73.7{\tiny$\pm$0.5}  & 75.3{\tiny$\pm$0.4}  & 76.9{\tiny$\pm$0.4}  & 76.9{\tiny$\pm$0.4}  & 76.8{\tiny$\pm$0.4}  & 67.3{\tiny$\pm$0.4}  & 70.5{\tiny$\pm$0.3}  & 73.2{\tiny$\pm$0.3}  & 73.1{\tiny$\pm$0.3}  & 72.9{\tiny$\pm$0.3} 
\\ 
& DPN$^{+}$ 
& 78.8{\tiny$\pm$0.1}  & 81.3{\tiny$\pm$0.1}  & 83.6{\tiny$\pm$0.1}  & 83.6{\tiny$\pm$0.1}  & 84.8{\tiny$\pm$0.1}  & 68.5{\tiny$\pm$0.2}  & 71.4{\tiny$\pm$0.1}  & 71.8{\tiny$\pm$0.1}  & 71.8{\tiny$\pm$0.1}  & 73.3{\tiny$\pm$0.1} 
\\
& DPN$^{-}$ 
& 77.5{\tiny$\pm$0.0}  & 79.5{\tiny$\pm$0.0}  & \textbf{85.2{\tiny$\pm$0.1}}  & \textbf{85.2{\tiny$\pm$0.1}}  & 58.4{\tiny$\pm$0.2}  & 71.9{\tiny$\pm$0.1}  & 74.4{\tiny$\pm$0.1}  & \textbf{78.9{\tiny$\pm$0.2}}  & \textbf{78.9{\tiny$\pm$0.2}}  & 52.1{\tiny$\pm$0.2} 
\\ 
\hline
\end{tabular}
}
\end{table*}

\begin{table*}[ht]
\centering
\caption{ Results of OOD image detection for TIM. 
	We report (mean $\pm$ standard deviation) of three different models.
	Description of these OOD datasets are provided in Appendix \ref{app:ood_description}.}
\label{table:extended_result_tim}
\vspace*{0.5em}
\resizebox{14.0cm}{!}{%
\begin{tabular}{ll|ccccc|ccccc}
\hline
\multirow{2}{*}{} & \multirow{2}{*}{Methods} & \multicolumn{5}{c|}{AUROC} & \multicolumn{5}{c}{AUPR} \\
\multicolumn{2}{c|}{} & Max.P & Ent. & MI & $\alpha_0$ & D-Ent & Max.P & Ent. & MI & $\alpha_0$ & D-Ent \\ \hline
\\
\multirow{6}{*}{\rotatebox{90}{CIFAR-10} } 
& Baseline 
& 76.9{\tiny$\pm$0.2}  & 79.2{\tiny$\pm$0.2}  & -  & -  & -  & 72.6{\tiny$\pm$0.2}  & 75.1{\tiny$\pm$0.1}  & -  & - & -  \\
& MCDP 
& 77.4{\tiny$\pm$0.1}  & 79.8{\tiny$\pm$0.1}  & 77.5{\tiny$\pm$0.2} &- &- & 73.0{\tiny$\pm$0.2}  & 75.4{\tiny$\pm$0.1}  & 71.4{\tiny$\pm$0.4}  & - & - \\
& DE 
& 76.9{\tiny$\pm$NA} & 79.3{\tiny$\pm$NA} & 77.7{\tiny$\pm$NA} &- &- & 72.6{\tiny$\pm$NA} & 75.2{\tiny$\pm$NA} & 72.4{\tiny$\pm$NA} &- &-
\\
& EDL 
& 74.4{\tiny$\pm$0.7}  & 76.1{\tiny$\pm$0.7}  & 72.0{\tiny$\pm$1.3}  & 71.8{\tiny$\pm$1.3}  & 73.3{\tiny$\pm$1.1}  & 69.4{\tiny$\pm$0.3}  & 70.9{\tiny$\pm$0.8}  & 67.8{\tiny$\pm$1.3}  & 67.7{\tiny$\pm$1.4}  & 68.7{\tiny$\pm$1.3}
\\ 
& OE 
& 91.3{\tiny$\pm$0.4}  & 92.6{\tiny$\pm$0.3}  & - & - & - & 92.9{\tiny$\pm$0.4}  & 94.0{\tiny$\pm$0.3}  & -  & -  & - \\ 
& DPN$_{rev}$ 
& 85.4{\tiny$\pm$0.7}  & 86.0{\tiny$\pm$0.8}  & 82.8{\tiny$\pm$1.4}  & 81.9{\tiny$\pm$1.6}  & 85.6{\tiny$\pm$0.9}  & 88.1{\tiny$\pm$0.7}  & 88.7{\tiny$\pm$0.7}  & 86.5{\tiny$\pm$1.3}  & 86.0{\tiny$\pm$1.4}  & 88.7{\tiny$\pm$0.8} \\ 

& DPN$^+$ 
& 99.2{\tiny$\pm$0.0}  & 99.4{\tiny$\pm$0.0}  & 99.7{\tiny$\pm$0.0}  & 99.7{\tiny$\pm$0.0}  & 99.6{\tiny$\pm$0.0}  & 99.4{\tiny$\pm$0.0}  & 99.5{\tiny$\pm$0.0}  & 99.7{\tiny$\pm$0.0}  & 99.7{\tiny$\pm$0.0}  & 99.6{\tiny$\pm$0.0} \\
& DPN$^-$  
& 99.7{\tiny$\pm$0.0}  & 99.8{\tiny$\pm$0.0}  & \textbf{99.9{\tiny$\pm$0.0}}  & \textbf{99.9{\tiny$\pm$0.0}}  & 3.5{\tiny$\pm$0.1}  & 99.8{\tiny$\pm$0.0}  & 99.8{\tiny$\pm$0.0}  & \textbf{99.9{\tiny$\pm$0.0}}  & \textbf{99.9{\tiny$\pm$0.0}}  & 33.6{\tiny$\pm$0.1}  \\
\hline
\\
\multirow{6}{*}{\rotatebox{90}{CIFAR-100} } 
& Baseline 
& 73.6{\tiny$\pm$0.2}  & 75.7{\tiny$\pm$0.2}  & - & -  & - & 69.7{\tiny$\pm$0.2}  & 72.2{\tiny$\pm$0.2}  & - & - & - \\
& MCDP  
& 74.0{\tiny$\pm$0.2}  & 76.1{\tiny$\pm$0.2}  & 73.6{\tiny$\pm$0.2}  &- &- & 70.0{\tiny$\pm$0.1}  & 72.3{\tiny$\pm$0.1}  & 67.6{\tiny$\pm$0.3} & - & - \\
& DE 
& 73.7{\tiny$\pm$NA} & 75.8{\tiny$\pm$NA} & 75.3{\tiny$\pm$NA} &- &- 
& 69.7{\tiny$\pm$NA} & 72.3{\tiny$\pm$NA} & 70.6{\tiny$\pm$NA} & - &-
\\
& EDL 
& 71.8{\tiny$\pm$0.6}  & 73.7{\tiny$\pm$0.7}  & 70.1{\tiny$\pm$0.9}  & 69.9{\tiny$\pm$0.9}  & 71.4{\tiny$\pm$0.8}  & 67.2{\tiny$\pm$0.4}  & 68.6{\tiny$\pm$0.9}  & 65.9{\tiny$\pm$1.0}  & 65.8{\tiny$\pm$1.0}  & 66.7{\tiny$\pm$1.0}
\\
& OE 
& 89.5{\tiny$\pm$0.5}  & 90.9{\tiny$\pm$0.4}  &- &-  &- & 91.5{\tiny$\pm$0.5}  & 92.7{\tiny$\pm$0.4}  &-  &-  &-  \\ 
& DPN$_{rev}$ 
& 84.2{\tiny$\pm$0.8}  & 85.0{\tiny$\pm$0.8}  & 82.5{\tiny$\pm$1.4}  & 81.7{\tiny$\pm$1.6}  & 85.0{\tiny$\pm$0.9}  & 87.0{\tiny$\pm$0.7}  & 87.9{\tiny$\pm$0.7}  & 86.3{\tiny$\pm$1.3}  & 85.7{\tiny$\pm$1.4}  & 88.2{\tiny$\pm$0.8}  \\ 

& DPN$^+$ 
& 98.8{\tiny$\pm$0.0}  & 99.0{\tiny$\pm$0.0}  & 99.5{\tiny$\pm$0.0}  & 99.5{\tiny$\pm$0.0}  & 99.4{\tiny$\pm$0.0}  & 99.1{\tiny$\pm$0.0}  & 99.3{\tiny$\pm$0.0}  & 99.5{\tiny$\pm$0.0}  & 99.5{\tiny$\pm$0.0}  & 99.5{\tiny$\pm$0.0}  \\

& DPN$^-$ 
& 98.7{\tiny$\pm$0.1}  & 99.0{\tiny$\pm$0.0}  & \textbf{99.6{\tiny$\pm$0.0}}  & \textbf{99.6{\tiny$\pm$0.0}}  & 7.5{\tiny$\pm$0.2}  & 99.1{\tiny$\pm$0.0}  & 99.3{\tiny$\pm$0.0}  & \textbf{99.7{\tiny$\pm$0.0}}  & \textbf{99.7{\tiny$\pm$0.0}}  & 36.7{\tiny$\pm$0.2}  \\
\hline
\\
\multirow{6}{*}{\rotatebox{90}{STL-10} } 
& Baseline 
& 67.7{\tiny$\pm$0.1}  & 67.7{\tiny$\pm$0.1}  & - & - & - & 56.5{\tiny$\pm$0.1}  & 56.8{\tiny$\pm$0.2}  & - & - & - \\
& MCDP 
& 68.1{\tiny$\pm$0.1}  & 68.1{\tiny$\pm$0.1}  & 69.8{\tiny$\pm$0.1} &- &- & 56.8{\tiny$\pm$0.2}  & 57.3{\tiny$\pm$0.2}  & 60.1{\tiny$\pm$0.4} & - & - \\
& DE 
& 67.7{\tiny$\pm$NA} & 67.8{\tiny$\pm$NA} & 68.1{\tiny$\pm$NA} &- &- & 56.5{\tiny$\pm$NA} & 56.8{\tiny$\pm$NA} & 58.1{\tiny$\pm$NA} & - &-
\\
& EDL
& 66.5{\tiny$\pm$0.3}  & 65.0{\tiny$\pm$0.6}  & 60.4{\tiny$\pm$1.0}  & 60.3{\tiny$\pm$1.0}  & 61.3{\tiny$\pm$0.9}  & 55.7{\tiny$\pm$0.1}  & 55.6{\tiny$\pm$0.4}  & 53.7{\tiny$\pm$0.7}  & 53.6{\tiny$\pm$0.8}  & 54.1{\tiny$\pm$0.7}
\\
& OE 
& \textbf{100.0{\tiny$\pm$0.0}}  & \textbf{100.0{\tiny$\pm$0.0}}  & - & - & - & \textbf{100.0{\tiny$\pm$0.0}}  & \textbf{100.0{\tiny$\pm$0.0}}  & - &- & - \\ 
& DPN$_{rev}$ 
& 99.4{\tiny$\pm$0.0}  & 99.5{\tiny$\pm$0.0}  & 99.6{\tiny$\pm$0.0}  & 99.6{\tiny$\pm$0.0}  & 99.5{\tiny$\pm$0.0}  & 99.5{\tiny$\pm$0.0}  & 99.6{\tiny$\pm$0.0}  & 99.7{\tiny$\pm$0.0}  & 99.7{\tiny$\pm$0.0}  & 99.6{\tiny$\pm$0.0} \\ 
& DPN$^+$ 
& 99.9{\tiny$\pm$0.0}  & \textbf{100.0{\tiny$\pm$0.0}}  & \textbf{100.0{\tiny$\pm$0.0}}  & \textbf{100.0{\tiny$\pm$0.0}}  & \textbf{100.0{\tiny$\pm$0.0}}  & 99.9{\tiny$\pm$0.0}  & \textbf{100.0{\tiny$\pm$0.0}}  & 99.9{\tiny$\pm$0.1}  & 99.9{\tiny$\pm$0.1}  & 99.9{\tiny$\pm$0.0} \\
& DPN$^-$ 
& \textbf{100.0{\tiny$\pm$0.0}}  & \textbf{100.0{\tiny$\pm$0.0}}  & \textbf{100.0{\tiny$\pm$0.0}}  & \textbf{100.0{\tiny$\pm$0.0}}  & 0.2{\tiny$\pm$0.0}  & \textbf{100.0{\tiny$\pm$0.0}} & \textbf{100.0{\tiny$\pm$0.0}}  & \textbf{100.0{\tiny$\pm$0.0}}  & \textbf{100.0{\tiny$\pm$0.0}}  & 26.6{\tiny$\pm$0.0}  \\ \hline
\\
\multirow{6}{*}{\rotatebox{90}{LSUN} } 
& Baseline 
& 69.0{\tiny$\pm$0.0}  & 70.2{\tiny$\pm$0.1}  & - & - & - & 63.5{\tiny$\pm$0.2}  & 64.0{\tiny$\pm$0.3}  & - & - & - \\
& MCDP 
& 69.3{\tiny$\pm$0.1}  & 70.4{\tiny$\pm$0.1}  & 69.5{\tiny$\pm$0.2}  &- &- & 63.6{\tiny$\pm$0.2}  & 64.0{\tiny$\pm$0.3}  & 63.2{\tiny$\pm$0.4}  & - & -\\
& DE 
& 69.1{\tiny$\pm$NA} & 70.2{\tiny$\pm$NA} & 70.3{\tiny$\pm$NA} &- &- 
& 63.5{\tiny$\pm$NA} & 64.0{\tiny$\pm$NA} & 65.2{\tiny$\pm$NA} &- &-
\\
& EDL
& 65.5{\tiny$\pm$0.2}  & 67.1{\tiny$\pm$0.6}  & 64.7{\tiny$\pm$1.3}  & 64.7{\tiny$\pm$1.3}  & 65.6{\tiny$\pm$1.2}  & 60.4{\tiny$\pm$0.3}  & 61.8{\tiny$\pm$0.8}  & 60.6{\tiny$\pm$1.5}  & 60.5{\tiny$\pm$1.6}  & 61.0{\tiny$\pm$1.4}
 \\
& OE 
& \textbf{100.0{\tiny$\pm$0.0}}  & \textbf{100.0{\tiny$\pm$0.0}}  & -  & -  & - & \textbf{100.0{\tiny$\pm$0.0}}  & \textbf{100.0{\tiny$\pm$0.0} } & -  & - & - \\ 
& DPN$_{rev}$ 
& 99.6{\tiny$\pm$0.0}  & 99.7{\tiny$\pm$0.0}  & 99.8{\tiny$\pm$0.0}  & 99.8{\tiny$\pm$0.0}  & 99.8{\tiny$\pm$0.0}  & 99.7{\tiny$\pm$0.0}  & 99.8{\tiny$\pm$0.0}  & 99.8{\tiny$\pm$0.0}  & 99.8{\tiny$\pm$0.0}  & 99.8{\tiny$\pm$0.0}  \\ 

& DPN$^+$ 
& \textbf{100.0{\tiny$\pm$0.0}}  & \textbf{100.0{\tiny$\pm$0.0}}  & \textbf{100.0{\tiny$\pm$0.0}}  & \textbf{100.0{\tiny$\pm$0.0}}  & \textbf{100.0{\tiny$\pm$0.0}}  & \textbf{100.0{\tiny$\pm$0.0}}  & \textbf{100.0{\tiny$\pm$0.0}}  & 99.9{\tiny$\pm$0.0}  & 99.9{\tiny$\pm$0.0}  & 99.9{\tiny$\pm$0.0} \\

& DPN$^-$ 
& \textbf{100.0{\tiny$\pm$0.0}}  & \textbf{100.0{\tiny$\pm$0.0}}  & \textbf{100.0{\tiny$\pm$0.0}}  & \textbf{100.0{\tiny$\pm$0.0}}  & 0.2{\tiny$\pm$0.0}  & \textbf{100.0{\tiny$\pm$0.0}}  & \textbf{100.0{\tiny$\pm$0.0}}  & \textbf{100.0{\tiny$\pm$0.0}}  & \textbf{100.0{\tiny$\pm$0.0}}  & 30.8{\tiny$\pm$0.0}  \\
\hline
\\
\multirow{6}{*}{\rotatebox{90}{Places365} } 
& Baseline 
& 71.1{\tiny$\pm$0.1}  & 72.7{\tiny$\pm$0.1}  & - & - & - & 87.5{\tiny$\pm$0.0}  & 88.2{\tiny$\pm$0.0}  & - & - & - \\
& MCDP 
& 71.4{\tiny$\pm$0.1}  & 73.0{\tiny$\pm$0.0}  & 70.8{\tiny$\pm$0.1} &- &- & 87.7{\tiny$\pm$0.0}  & 88.2{\tiny$\pm$0.0}  & 86.8{\tiny$\pm$0.2}  & - & - \\
& DE 
& 71.2{\tiny$\pm$NA} & 72.7{\tiny$\pm$NA} & 71.6{\tiny$\pm$NA} &- &- 
& 87.6{\tiny$\pm$NA} & 88.2{\tiny$\pm$NA} & 87.6{\tiny$\pm$NA} &- &-
\\
& EDL 
& 69.0{\tiny$\pm$0.1}  & 70.5{\tiny$\pm$0.2}  & 67.1{\tiny$\pm$0.4}  & 66.9{\tiny$\pm$0.4}  & 68.2{\tiny$\pm$0.4}  & 86.5{\tiny$\pm$0.2}  & 87.1{\tiny$\pm$0.2}  & 86.0{\tiny$\pm$0.3}  & 86.0{\tiny$\pm$0.3}  & 86.4{\tiny$\pm$0.3}
\\
& OE 
& 99.9{\tiny$\pm$0.0}  & 99.9{\tiny$\pm$0.0}  & - & -  & -  & \textbf{100.0{\tiny$\pm$0.0}}  & \textbf{100.0{\tiny$\pm$0.0}}  & -  & - & - \\ 
& DPN$_{rev}$ 
& 98.9{\tiny$\pm$0.1}  & 99.1{\tiny$\pm$0.0}  & 99.2{\tiny$\pm$0.1}  & 99.2{\tiny$\pm$0.1}  & 99.2{\tiny$\pm$0.0}  & 99.7{\tiny$\pm$0.0}  & 99.8{\tiny$\pm$0.0}  & 99.8{\tiny$\pm$0.0}  & 99.8{\tiny$\pm$0.0}  & 99.8{\tiny$\pm$0.0}  \\ 
& DPN$^+$  
& 99.9{\tiny$\pm$0.0}  & 99.9{\tiny$\pm$0.0}  & 99.9{\tiny$\pm$0.0}  & 99.9{\tiny$\pm$0.0}  & 99.9{\tiny$\pm$0.0}  & \textbf{100.0{\tiny$\pm$0.0}} & \textbf{100.0{\tiny$\pm$0.0} } & \textbf{100.0{\tiny$\pm$0.0}}  & \textbf{100.0{\tiny$\pm$0.0}}  & \textbf{100.0{\tiny$\pm$0.0}} \\

& DPN$^-$ 
& 99.9{\tiny$\pm$0.0}  & 99.9{\tiny$\pm$0.0}  & \textbf{100.0{\tiny$\pm$0.0}}  & \textbf{100.0{\tiny$\pm$0.0}}  & 0.8{\tiny$\pm$0.0}  & \textbf{100.0{\tiny$\pm$0.0}}  & \textbf{100.0{\tiny$\pm$0.0}}  & \textbf{100.0{\tiny$\pm$0.0}}  & \textbf{100.0{\tiny$\pm$0.0}}  & 58.5{\tiny$\pm$0.0}  \\
\hline
\\
\multirow{6}{*}{\rotatebox{90}{Textures} } 
& Baseline 
& 70.9{\tiny$\pm$0.2}  & 73.3{\tiny$\pm$0.2}  & - & - & -  & 53.8{\tiny$\pm$0.4}  & 57.5{\tiny$\pm$0.5}  & -  & - & - \\
& MCDP 
& 70.3{\tiny$\pm$0.2}  & 72.6{\tiny$\pm$0.3}  & 63.6{\tiny$\pm$0.2}  &- &- & 53.1{\tiny$\pm$0.3}  & 56.5{\tiny$\pm$0.5}  & 41.6{\tiny$\pm$0.2}  & - & - \\
& DE
& 71.1{\tiny$\pm$NA} & 73.4{\tiny$\pm$NA} & 76.2{\tiny$\pm$NA} &- &- & 53.9{\tiny$\pm$NA} & 57.7{\tiny$\pm$NA} & 62.2{\tiny$\pm$NA} &- &-
\\
& EDL 
& 71.0{\tiny$\pm$0.1}  & 74.0{\tiny$\pm$0.4}  & 70.4{\tiny$\pm$0.9}  & 70.2{\tiny$\pm$0.9}  & 71.9{\tiny$\pm$0.9}  & 52.8{\tiny$\pm$0.5}  & 55.8{\tiny$\pm$0.4}  & 54.3{\tiny$\pm$0.9}  & 54.2{\tiny$\pm$0.9}  & 54.9{\tiny$\pm$0.8}
\\
& OE 
& 95.8{\tiny$\pm$0.3}  & 96.4{\tiny$\pm$0.3}  & - & - & -  & 95.5{\tiny$\pm$0.3}  & 96.1{\tiny$\pm$0.3}  & -  & -  & - \\ 
& DPN$_{rev}$ 
& 90.9{\tiny$\pm$0.3}  & 91.9{\tiny$\pm$0.3}  & 91.2{\tiny$\pm$0.6}  & 90.6{\tiny$\pm$0.6}  & 92.6{\tiny$\pm$0.3}  & 89.9{\tiny$\pm$0.3}  & 91.3{\tiny$\pm$0.3}  & 90.7{\tiny$\pm$0.6}  & 90.2{\tiny$\pm$0.6}  & 92.3{\tiny$\pm$0.3} \\ 

& DPN$^+$  
& 96.5{\tiny$\pm$0.1}  & 97.1{\tiny$\pm$0.0}  & 98.4{\tiny$\pm$0.0}  & 98.4{\tiny$\pm$0.0}  & 98.2{\tiny$\pm$0.0}  & 96.2{\tiny$\pm$0.0}  & 96.9{\tiny$\pm$0.0}  & 97.9{\tiny$\pm$0.0}  & 97.9{\tiny$\pm$0.0}  & 97.8{\tiny$\pm$0.0}  \\

& DPN$^-$  
& 95.8{\tiny$\pm$0.1}  & 96.8{\tiny$\pm$0.1}  & \textbf{98.7{\tiny$\pm$0.1}}  & \textbf{98.7{\tiny$\pm$0.1}}  & 19.3{\tiny$\pm$0.4}  & 95.4{\tiny$\pm$0.0}  & 96.5{\tiny$\pm$0.0}  & \textbf{98.5{\tiny$\pm$0.1}}  & \textbf{98.5{\tiny$\pm$0.1}}  & 35.7{\tiny$\pm$0.3} \\
\hline
\end{tabular}
}
\end{table*}

\end{document}